\documentclass[10pt,journal,compsoc]{IEEEtran}

\usepackage{amsmath}
\usepackage{color}
\usepackage{colortbl}
\usepackage[pagebackref=true,breaklinks=true,colorlinks,bookmarks=false]{hyperref}
\usepackage{amssymb}
\usepackage{longtable}
\usepackage{xcolor}
\usepackage{siunitx}
\usepackage{pbox}
\usepackage{multirow}
\usepackage{times}

\ifCLASSOPTIONcompsoc
\usepackage[nocompress]{cite}
\else
\usepackage{cite}
\fi

\ifCLASSINFOpdf
\usepackage[pdftex]{graphicx}
\else
\fi

\ifCLASSOPTIONcompsoc
\usepackage[caption=false,font=footnotesize,labelfont=sf,textfont=sf]{subfig}
\else
\usepackage[caption=false,font=footnotesize]{subfig}
\fi

\usepackage{romannum}

\begin{document}
\pagenumbering{arabic}
\title{Bottom-Up Human Pose
Estimation by 
Ranking
Heatmap-Guided Adaptive Keypoint Estimates}

\author{
Ke Sun,
Zigang Geng,
Depu Meng,
Bin Xiao,
Dong Liu, 
Zhaoxiang Zhang,
Jingdong Wang
\IEEEcompsocitemizethanks{\IEEEcompsocthanksitem J. Wang is with Microsoft Research,
Beijing, P.R. China.\protect\\
E-mail: jingdw@microsoft.com
}%
}

\markboth{Submitted to IEEE Transactions on Pattern Analysis and Machine Intelligence, June~2020}%
{Wang \MakeLowercase{\textit{et al.}}: Bottom-Up Human Pose Estimation with Heatmap-Guided Adaptive Keypoint Regressionand Shape Scoring}

\IEEEtitleabstractindextext{%
\begin{abstract}
The typical bottom-up human pose estimation
framework includes
two stages,
keypoint detection
and grouping.
Most existing works focus on 
developing grouping algorithms,
e.g.,
associative embedding, and pixel-wise keypoint regression 
that we adopt in our approach.
We present several schemes that
are rarely or unthoroughly studied before
for improving keypoint detection and grouping
(keypoint regression) performance.
First,
we exploit the keypoint heatmaps 
for pixel-wise keypoint regression
instead of separating them\footnote{The separation means that the heads
	for heatmap estimation and pixel-wise keypoint regression
	are separate, but they might share the same backbone.}
for improving keypoint regression.
Second, we adopt a pixel-wise spatial transformer network to learn adaptive representations
for handling the scale and orientation variance
to further improve keypoint regression quality.
Last, we present a joint shape and heatvalue scoring scheme
to promote the estimated poses that are more likely to be true poses.
Together with the tradeoff heatmap estimation loss for 
balancing the background and keypoint pixels
and thus improving heatmap estimation quality, 
we get the state-of-the-art bottom-up human pose estimation result.
Code is available at https://github.com/HRNet/HRNet-Bottom-up-Pose-Estimation.
\end{abstract}

}

\maketitle

\IEEEdisplaynontitleabstractindextext

\IEEEpeerreviewmaketitle

\ifCLASSOPTIONcompsoc
\IEEEraisesectionheading{\section{Introduction}\label{sec:introduction}}
\else
\section{Introduction}
\label{sec:introduction}
\fi

\begin{figure*}[t]
	\small
	\centering
	\includegraphics[width=0.24\textwidth, height = 0.185\textwidth]{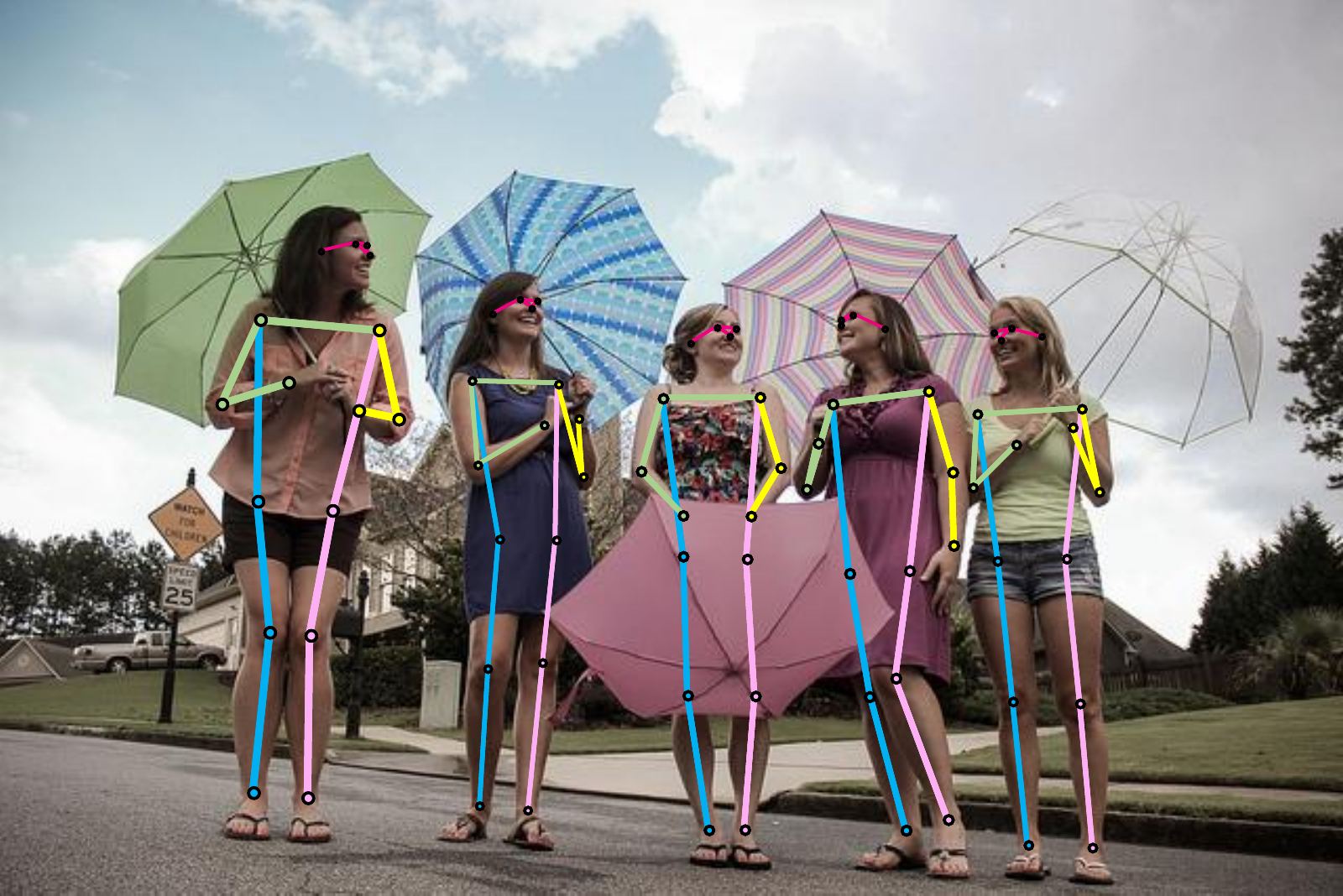}
	\includegraphics[width=0.24\textwidth, height = 0.185\textwidth]{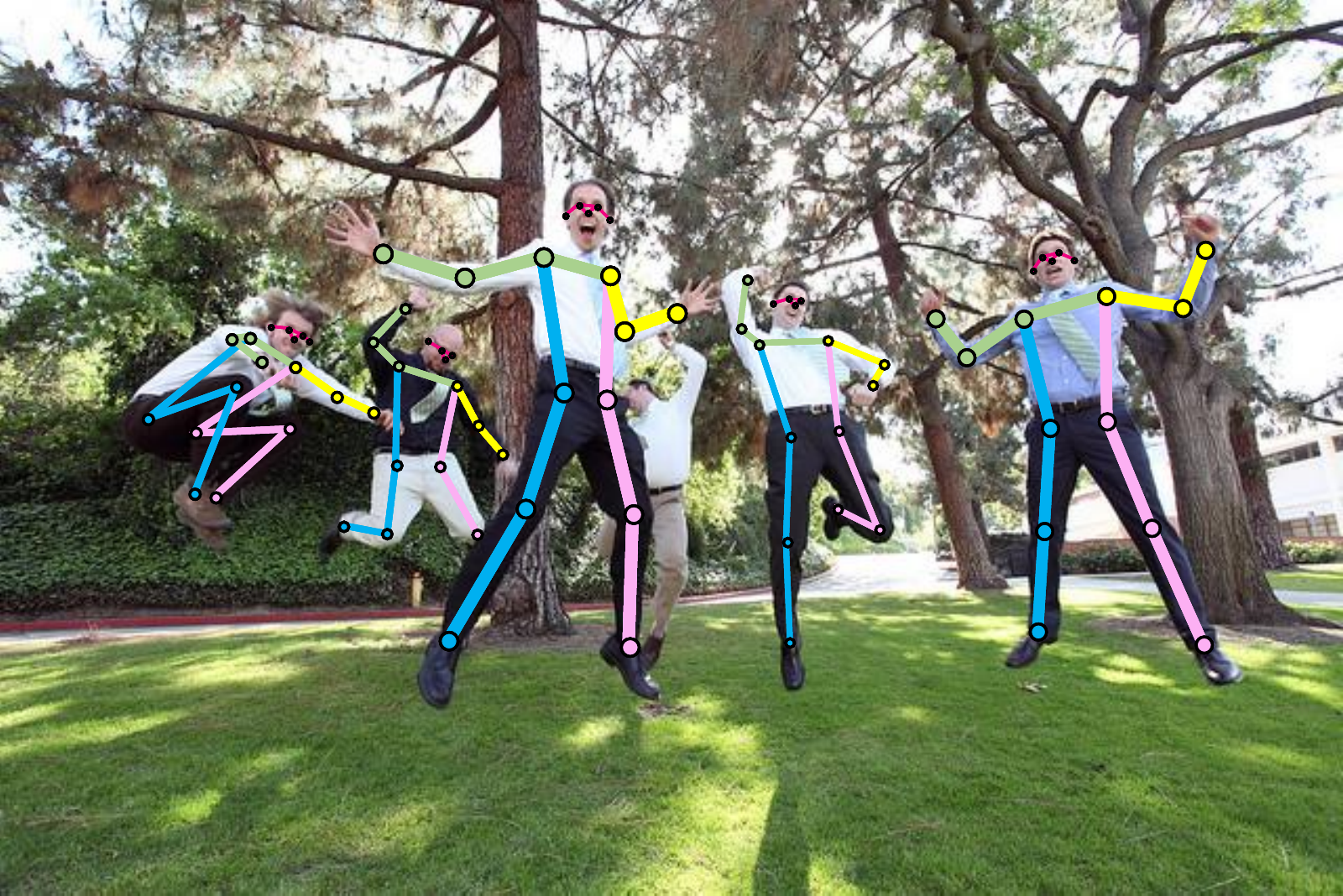}
	\includegraphics[width=0.24\textwidth, height = 0.185\textwidth]{fig/coco_examples/cocoid_13291.pdf}
	\includegraphics[width=0.24\textwidth, height = 0.185\textwidth]{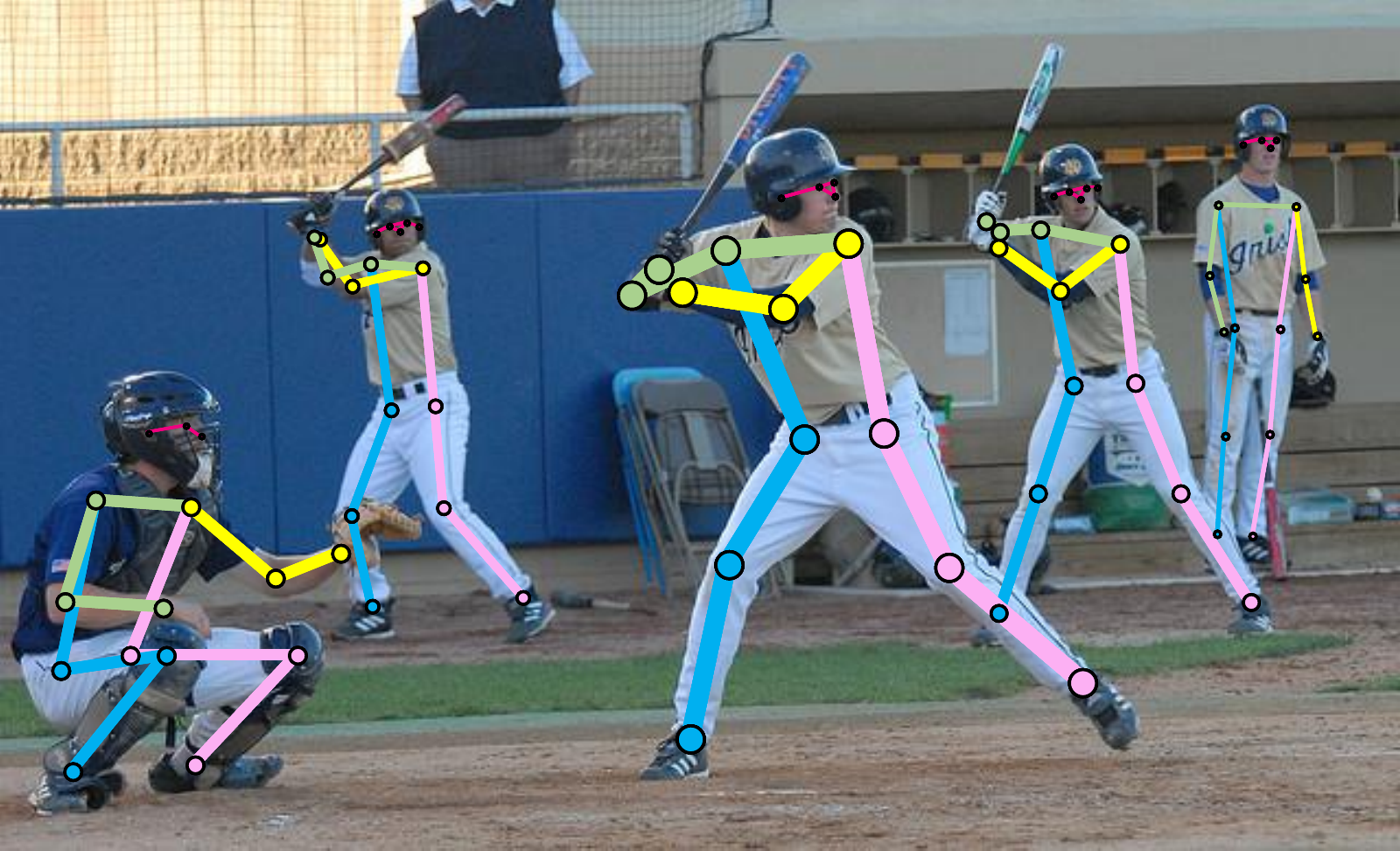}
	\vspace{-.3cm}
	\caption{Multi-person pose estimation.
		There are several key challenges for the bottom-up framework:
		unknown number of persons,
		diverse person scales
		and orientations,
		various poses,
		etc.}
	\label{fig:teaser}
\end{figure*}

\IEEEPARstart{H}{uman} pose estimation
aims to predict the keypoint positions
of each person from an image,
i.e., localize the keypoints as well as identify the person the keypoints belong to.
It has broad applications, 
including action recognition, 
person  re-identification,
pedestrian tracking, human-computer interaction, smart photo editing, etc.
A lot of techniques have been developed
to deal with various challenges as depicted in Figure~\ref{fig:teaser},
such as unknown number of persons,
diverse person scales
and orientations,
various poses, and so on.

There are two main frameworks 
top-down and bottom-up.
The top-down framework first detects the person
and then performs single-person pose estimation
for each detected person.
The bottom-up framework first predicts the keypoint positions 
and then groups the keypoints into individuals.
The former is more accurate
but more costly,
and the latter is more efficient and less accurate.
This paper focuses on the latter one 
and improving the pose estimation accuracy.

The typical bottom-up framework consists of two steps.
The first step is to estimate keypoint heatmaps,
where each position in each keypoint heatmap has a value
indicating the degree that the keypoint lies in the position. 
The second step is to group the detected keypoints,
identified in the first step,
into persons.
Most existing works mainly focus on the second step.
The representative works include affinity linking~\cite{CaoSWS17,Hidalgo_2019_ICCV},
associative embedding~\cite{NewellHD17},
pixel-wise keypoint regression~\cite{ZhouWK19},
and so on.
We adopt the pixel-wise keypoint regression scheme,
where the pose is represented
by a center point and the offsets for each keypoint
to the center point,
and regard the regression results
as the grouping cues to absorb the keypoints detected from the heatmaps,
with the focus on
improving the heatmap estimation and pixel-wise keypoint regression quality.

We combine the predicted keypoint heatmaps
and the feature representation
for pixel-wise keypoint regression.
It differs from previous schemes that separate
the heatmap estimation head and
the pixel-wise keypoint regression head~\cite{ZhouWK19}.
As a result,
the heatmaps, 
whose quality is usually better 
than pixel-wise keypoint regression 
in terms of keypoint localization quality,
provide strong guide to pixel-wise keypoint regression,
leading higher keypoint regression quality.

We adopt a pixel-wise spatial transformer network,
a simple extension of spatial transformer network~\cite{JaderbergSZK15},
to adaptively learn the representation
for handling the local transformation variance,
such as human scale and orientation variance.
This is motivated by that pixel-wise keypoint regression 
is an object-level (person) task
and different persons in one image might 
have different scales
and/or different orientations.

In addition,
we present a joint shape and heatvalue scoring scheme to predict the degree 
that each pose estimation
is a real pose.
We use the scores to rank 
the final pose estimation results
by demoting the mis-grouped poses
(e.g., keypoints absorbed together
but from different persons or the background).
We also revisit the imbalance issue~\cite{PishchulinITAAG16}
between keypoint pixels and non-keypoint pixels,
and simply reweigh the two kinds of pixels
in the heatmap estimation loss,
which improves the heatmap estimation quality significantly.
We demonstrate the proposed approach
with the state-of-the-art bottom-up human pose estimation performance on the COCO and CrowdPose benchmark.
We obtain the AP score $70.2$ for the single-scale testing on the COCO test-dev set and the AP score $66.2$ on the CrowdPose test set.

\section{Related Work}
The convolutional neural network solutions~\cite{GkioxariTJ16,LifshitzFU16,TangYW18,NieFY18,NieFZY18,PengTYFM18,SunLXZLW17,FanZLW15} 
	to human pose estimation
	have shown superior performance
	over the conventional methods,
	such as the probabilistic graphical model
	or the pictorial structure model~\cite{YangR11,PishchulinAGS13}.
	Recent advances show that 
	the heatmap estimation based methods,
	estimating keypoint heatmaps~\cite{ChuOLW16,ChuYOMYW17,YangOLW16}
	where the keypoints are localized,
	outperform the keypoint position prediction methods~\cite{ToshevS14,CarreiraAFM16}.

	\vspace{.1cm}
	\noindent\textbf{Top-down methods.}
	Representative works include:
	PoseNet~\cite{PapandreouZKTTB17}, 
	RMPE~\cite{FangXTL17}, 
	convolutional pose machine~\cite{WeiRKS16},
	Hourglass~\cite{NewellYD16},
	Mask R-CNN~\cite{HeGDG17}, and cascaded pose networks~\cite{ChenWPZYS18},
	simple baseline~\cite{XiaoWW18},
	and so on.
	The recently-developed HRNet~\cite{SunXLW19, Wang2020}
	achieves the significant gain,
	especially regarding keypoint localization accuracy.
	These methods exploit the
	advances in person detection 
	as well as extra person bounding-box labeling information~\cite{KreissBA19}.
	The keypoint heatmap estimation 
	is eased 
	as the background is largely removed
	and fewer confusing pixels are remained
	(in most cases there is only one instance
	for each keypoint)
	in the detection box.
	The top-down pipeline, however,
	takes extra cost 
	in the person box detection.
	
	\vspace{.1cm}
	\noindent\textbf{Bottom-up methods.}
	Most existing bottom-up methods mainly focus on 
	how to associate the detected keypoints 
	that belong to the same person
	together.
	The pioneering work,
	DeepCut~\cite{PishchulinITAAG16} and DeeperCut~\cite{InsafutdinovPAA16},
	formulates the keypoint association problem
	as an integer linear program,
	which however takes longer processing time
	(e.g., the order of hours).
	
	The OpenPose work~\cite{CaoSWS17}, 
	a real-time pose detector,
	developed 
	the part-affinity field approach 
	to link the keypoints that are likely to lie
	in the same person,
	which is extended in
	the PifPaf work~\cite{KreissBA19}.
	The associative embedding approach~\cite{NewellHD17}
	maps each keypoint to a scalar embedding
	so that the embeddings of the keypoints from the same person 
	are close,
	and clusters the keypoints using the scalar embeddings.
	The PersonLab approach~\cite{PapandreouZCGTM18}
	introduces a greedy decoding scheme
	together with hough voting for grouping.

	Several recent works~\cite{ZhouWK19,NieZYF19,NieFY18} 
	densely 
	regress a set of pose candidates,
	where each candidate consists of the keypoint positions
	that might be from the same person,
	and then use the candidates as grouping cues
	to cluster the keypoints selected
	from the keypoint heatmaps 
	into individuals.
	Our work belongs to this category
	and proposes
	to use heatmaps to
	guide pixel-wise keypoint regression.
	This guidance is
	in some sense related to 
	some other methods,
	e.g.,
	using part affinity fields
	to help predict heatmaps in OpenPose~\cite{CaoSWS17}.
	
	Additionally,
	we handle the scale and orientation diversity
	of different persons
	by a pixel-wise extension
	of spatial transformer network (STN)~\cite{JaderbergSZK15},
	pixel-wise STN.
	There are several related works, 
	dense spatial transformer network~\cite{LiCCDJ19}, 
	instance transformation network~\cite{WangZ0WT18},
	and deformable GANs~\cite{SiarohinSLS18}.
	The first one,
	dense spatial transformer network~\cite{LiCCDJ19},
	actually stills uses a global transformation (thin-plate spline).
	The second one is very close to ours,
	but applied to a different problem, text detection.
	The third one is to conduct STN over
	different regions
	for alignment,
	which comes from the pose estimation results,
	while ours is for pixel-level representations
	and thus robust pose estimation.
	The feature pyramid network~\cite{LinDGHHB17}
	and deformable convolutions~\cite{DaiQXLZHW17},
	mainly developed for object detection,
	might have similar effect
	with pixel-wise STN.
	We choose pixel-wise STN
	because 
	the implementation is easy without the necessity of distributing the objects
	of some scale into some pyramid level
	and it is more interpretable
	(explicitly model the scale and the orientation)
	compared to deformable convolutions.
	
	\begin{figure*}[t]
		\small
		\centering
		\includegraphics[width = 0.99\textwidth]{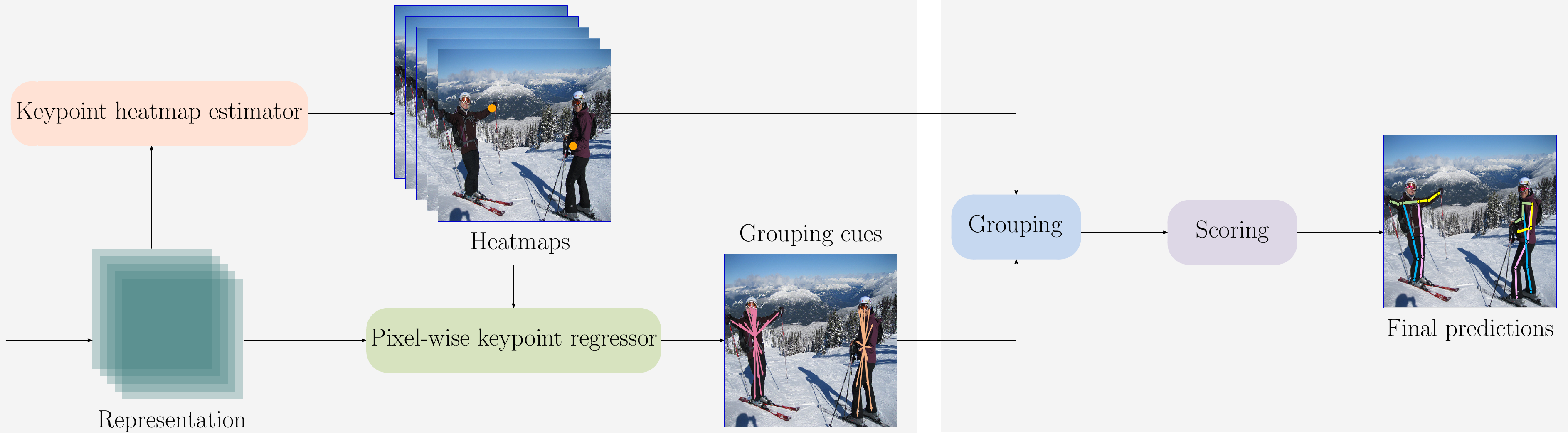}
		\vspace{-.3cm}
		\caption{
			Pipeline. The representation ($\mathsf{F}$) output from a backbone
			goes into the keypoint heatmap estimator $\mathcal{H}$, 
			outputting the keypoint heatmaps $\mathsf{H}$.
			The representation, the concatenation of the heatmaps $\mathsf{H}$
			and the input representation $\mathsf{F}$,
			is fed into the pixel-wise keypoint regressor $\mathcal{P}$,
			outputting the offset maps $\mathsf{O}$
			and the center heatmap $\mathbf{C}$
			as the grouping cues.
			The keypoint candidates obtained from the keypoint heatmaps $\mathsf{H}$
			are grouped with the help of the pixel-wise keypoint regression results
			$\mathsf{O}$ and $\mathbf{C}$,
			followed by a scoring stage
			generating the final pose predictions.}
		\label{fig:pipline}
	\end{figure*}
	
\section{Approach}
	Given an image $\mathsf{I}$,
	multi-person pose estimation 
	aims to
	predict a set of $N$ human poses:
	$\{\mathcal{P}_1, \mathcal{P}_2, \cdots, \mathcal{P}_N\}$,
	where the pose $\mathcal{P}_n=\{\mathbf{p}_{n1}, \mathbf{p}_{n2},
	\cdots, \mathbf{p}_{nK}\}$
	consists of $K$ keypoints belonging to the person $n$.

	\subsection{Formulation}
	The input image $\mathsf{I}$
	is fed into a backbone network,
	outputting a representation $\mathsf{F}$.
	The representation $\mathsf{F}$
	goes through the heads,
	as depicted in Figure~\ref{fig:pipline},
	with the outputs consisting of two parts:
	the keypoint heatmaps,
	and the pixel-wise keypoint regression results.
	The keypoint heatmaps $\mathsf{H}$
	consist of $K$ maps,
	$\mathbf{H}_1, \mathbf{H}_2, \cdots,
	\mathbf{H}_K$.
	The heat value at each position
	for each keypoint heatmap 
	indicates the degree
	that the keypoint lies in the position.
	
	The pixel-wise keypoint regression results
	consist of two parts.
	One is the center heatmap $\mathbf{C}$
	showing the degree that each position is the
	center of one pose.
	The offset maps, $\mathsf{O}$,
	contain $2K$ maps
	and 
	show the offsets
	of the keypoint to the center
	if the current position is a pose center.

	\vspace{.1cm}
	\noindent\textbf{Heatmap-guided pixel-wise keypoint regression.}
	\label{para:keypoint-estimation}
	The keypoint heatmaps are estimated from the representation
	through a keypoint estimation head $\mathcal{H}$:
	\begin{align}
	\mathsf{H} = \mathcal{H}(\mathsf{F}).
	\end{align}
	
	Unlike previous works~\cite{NieFXY18,NieZYF19}
	that regress the center heatmap $\mathbf{C}$
	and the pose coordinates $\mathbf{O}$ only from the representation $\mathsf{F}$,
	we estimate them by exploring the estimated heatmaps
	through a pixel-wise keypoint regression head $\mathcal{P}$,
	\begin{align}
	(\mathbf{C}, \mathsf{O}) = \mathcal{P}(\mathsf{F}, \mathsf{H}).
	\end{align}
	
	There are two benefits.
	The keypoint positions
	by the pixel-wise keypoint regression are regressed more accurately
	as the estimated keypoint heatmaps,
	relatively more accurate than pixel-wise keypoint regression,
	provide strong guidance for 
	regressing keypoint offsets. 
	On the other hand,
	heatmap estimation, besides the heatmap supervision,
	gets 
	additional supervision 
	from the pixel-wise keypoint regression target.

	\vspace{.1cm}
	\noindent\textbf{Adaptive representation transformation.}
	To address the scale and orientation variance,
	i.e., different persons in an image might 
	have different sizes and different orientations,
	we propose an adaptive representation transformation (ART) unit
	that consists of an adaptive convolution
	followed by BN and ReLU.
	The adaptive convolution
	is a modification of a normal convolution:
	$\mathbf{y}(\mathbf{q})= \sum\nolimits_{i=1}^9 \mathbf{W}_i \mathbf{x}(\mathbf{g}_{si}+ \mathbf{q}) $.
	Here, 
	$\mathbf{q}$ is a $2$D position,
	$\{\mathbf{g}_{s1}, \mathbf{g}_{s2}, 
	\dots, \mathbf{g}_{s9}\}$
	(denoted by a $2\times 9$ matrix $\mathbf{G}_s$)
	are $2$D offsets,
	and 
	$\{\mathbf{W}_1, \mathbf{W}_2, \dots,
	\mathbf{W}_9\}$ are the kernel weights.
	This modification is similar to deformable convolutions~\cite{DaiQXLZHW17},
	but we compute
	the offsets by explicitly modeling local scale and orientation.
	
	We compute the offsets $\mathbf{G}_s$:
	$\mathbf{G}_s = \mathbf{T}\mathbf{G}_t$,
	by estimating a local inverse affine transformation
	$\mathbf{T}$ ($\in \mathbb{R}^{2\times 2}$) 
	that characterizes local scaling and rotation
	for each position $\mathbf{q}$
	so that in the transformed space
	a convolution is conducted
	with the regular $3 \times 3$ positions,
	i.e., the offsets in a matrix form are:
	\makeatletter
	\renewcommand*\env@matrix[1][c]{\hskip -\arraycolsep
		\let\@ifnextchar\new@ifnextchar
		\array{*\c@MaxMatrixCols #1}}
	\makeatother
	\begin{align}
	\mathbf{G}_t = \begin{bmatrix}[r]
	-1 & ~0 & ~1 & -1 & ~~0 & ~~1 & -1 & ~~0 & ~~1 \\[0.3em]
	-1 & -1 & -1 & ~0 & ~~0 & ~~0 & ~1 & ~~1 & ~~1 
	\end{bmatrix}.
	\end{align}
	We estimate the local transformation $\mathbf{T}$
	by extending spatial transformer network~\cite{JaderbergSZK15}
	from a global manner
	to a pixel-wise manner,
	for each position through a $3\times3$ convolution 
	with the weights shared by all the positions.
	
	In our implementation,
	the keypoint heatmap estimation head $\mathcal{H}$
	consists of two adaptive representation transformation units
	and a subsequent linear transformation outputting the keypoint heatmaps $\mathsf{H}$.
	The pixel-wise keypoint regression head
	feeds the concatenation of 
	$\mathsf{F}$ and $\mathsf{H}$
	into two adaptive representation transformation units
	and a subsequent linear transformation outputting the center heatmap $\mathbf{C}$
	and the offset maps $\mathsf{O}$.
	
	\vspace{.1cm}
	\noindent\textbf{Loss function.}
	The loss contains two parts:
	keypoint heatmap estimation loss
	and 
	pixel-wise keypoint regression loss.
	We use a tradeoff heatmap estimation loss
	to balance the keypoint region
	and the non-keypoint region.
	The loss function is formulated 
	as the weighted distances
	between the predicted heat values
	and the groundtruth heat values:
	\begin{align}
	\ell_h = \| \mathsf{M} \odot (\mathsf{H} - \mathsf{H}^*)\|_2^2.
	\end{align}
	Here, $\|\cdot\|_2$ is the entry-wise $2$-norm.
	$\odot$ is the element-wise product operation.
	$\mathsf{M}$ corresponds $K$ masks,
	and the size is $H \times W \times K$.
	The $k$th mask, $\mathsf{M}_k$,
	is formed 
	so that the mask weight 
	of the positions not lying in the $k$th keypoint region is $0.1$,
	and others are $1$.
	
	We use the normalized smooth loss
	to form the pixel-wise keypoint regression loss:
	\begin{align}
	\ell_p = \sum_{i \in \mathcal{C}}
	\frac{1}{Z_i} \operatorname{smooth}_{L_1}(\mathbf{o}_i - \mathbf{o}^*_i)
	+ \|\mathbf{C} - \mathbf{C}^*\|_2^2.
	\end{align}
	Here, $Z_i = \sqrt{H_i^2+W_i^2}$ is the size
	of the corresponding person instance
	and $H_i$ and $W_i$ are the height and the width of 
	the instance box.
	$i$ is a position
	lying in a center region,
	and $\mathcal{C}$ is the set
	of the positions lying in some center region.
	$\mathbf{o}_i$ ($\mathbf{o}_i^*$),
	a column of the offset maps $\mathsf{O}$
	($\mathsf{O}^*$)
	is the $2K$-dimensional (groundtruth) offset vector 
	for the position $i$.
	$\mathbf{C}^*$
	is the groundtruth center heatmap.
	
	The whole loss function
	is the sum of the losses:
	\begin{align}
	\ell = \ell_h + \lambda \ell_p,
	\end{align}
	where $\lambda$
	is a weight.
	In our implementation,
	we consider that 
	the pixel-wise keypoint regression
	only provides a grouping cue,
	and
	set $\lambda = 0.01$,
	a smaller weight for the pixel-wise keypoint regression loss.

	\vspace{.1cm}
	\noindent\textbf{Training data construction.}
	We need to construct the training data
	for two parts: the keypoint heatmaps, 
	as well as
	the offset maps
	and the center heatmap.
	We generate the groundtruth keypoint heatmaps $\mathsf{H}^*$ for each training image.
	The groundtruth keypoint heatmaps $\mathsf{H}^*$ contains $K$ maps,
	and each map corresponds to one keypoint type.
	We build them
	as done in~\cite{NewellHD17}:
	assigning a heat value
	using the Gaussian function
	centered at a point
	around each groundtruth keypoint.
	
	The groundtruth offset maps and the groundtruth center heatmap are
	constructed
	from $\{\mathcal{P}_1, \mathcal{P}_2, \cdots, \mathcal{P}_N\}$.
	We use the $n$th pose $\mathcal{P}_n$ as an example and others are the same.
	We compute the center position $\bar{\mathbf{p}}_n = \frac{1}{K}\sum_{k=1}^K\mathbf{p}_{nk}$
	and the offsets
	$\mathcal{T}_n = \{\mathbf{p}_{n1} - \bar{\mathbf{p}}_n,
	\mathbf{p}_{n2} - \bar{\mathbf{p}}_n,
	\cdots,
	\mathbf{p}_{nK} - \bar{\mathbf{p}}_n
	\}$
	as the target.
	We use an expansion scheme
	to augment the center point
	to the center region:
	$\{\mathbf{m}_n^1, \mathbf{m}_n^2, \cdots, \mathbf{m}_n^M\}$, 
	which are central positions around 
	the pose center $\bar{\mathbf{p}}_n$
	with the radius $4$,
	and accordingly update the offsets.
	Each central position $\mathbf{m}_n^m$ has 
	a confidence value $c_n^m$
	indicating how confident it is the center
	and computed using the way forming the heatmap,
	which results in a so-called center heatmap $\mathbf{C}^*$\footnote{In case that one position belongs to two or more central regions, we choose only one central region whose center is the closest to that position.}.
	The positions not lying in the region
	have zero heat value, 
	and accordingly it have no offset values.
	The offset maps are denoted by $\mathsf{O}^*$.

	\subsection{Inference}
	\noindent\textbf{Grouping.}
	Given an image from which 
	we want to predict the human poses,
	we compute the keypoint heatmaps $\mathsf{H}$
	and the pixel-wise keypoint regression results $(\mathbf{C}, \mathsf{O})$.
	We use non-maximum suppression to find several ($30$ in our implementation) keypoint candidates,
	$\mathcal{S} = \{\mathcal{S}_1, \mathcal{S}_2, \dots, \mathcal{S}_K\}$,
	with each set $\mathcal{S}_k$
	consisting of the $k$th keypoint candidates
	from the keypoint heatmaps $\mathsf{H}$,
	and remove the keypoint candidates 
	whose heatvalues are too small (smaller than $0.01$).
	We also use non-maximum suppression
	to filter out pixel-wise keypoint regression results
	using the center heatmap,
	leading to a set of $M$ ($M = 30$, in our implementation) regression results,
	$\mathcal{G}_1, \dots, \mathcal{G}_M$,
	where each result $\mathcal{G}_m$
	contains $K$ keypoints.
	
	We group the keypoint candidates $\mathcal{S}$
	by regarding each pixel-wise keypoint regression result
	$\mathcal{G}_m$
	as a grouping cue.
	For each keypoint, e.g.,
	$\mathbf{p}_k$, in each group cue $\mathcal{G}_m$,
	we absorb 
	the closest keypoint candidate 
	among the candidates $\mathcal{S}_k$ with the same keypoint type
	if their distance is within $75$ pixels,
	and otherwise use $\mathbf{p}_k$ as the $k$th keypoint
	to form the $m$th pose candidate $\bar{\mathcal{G}}_m$.
	The $K$ keypoint candidates absorbed to $\mathcal{G}_m$ form a final pose candidate $\bar{\mathcal{G}}_m$.

	\vspace{.1cm}
	\noindent\textbf{Scoring.} 
	Given a candidate pose 
	$\bar{\mathcal{G}} = \{\mathbf{p}_1, \mathbf{p}_2, \cdots, \mathbf{p}_K\}$
	and the center point $\mathbf{p}$ predicting its grouping cue $\mathcal{G}$,
	the naive scoring scheme is 
	$
	\frac{1}{K}\sum_{k}h_k(\mathbf{p}_k)o(\mathbf{p})$,
	where $\mathbf{o}(\mathbf{p})$ is the heatvalue of $\mathbf{p}$
	from the center heatmap $\mathbf{O}$,
	and 
	$h_k(\mathbf{p}_k)$
	is the keypoint heatvalue from
	the $k$th keypoint heatmap $\mathbf{H}_k$.
	
	This naive scoring scheme does not consider
	the spatial information,
	and the space remains for improvement.
	Partially inspired by~\cite{RonchiP17},
	which suggests using a graphical model to capture the spatial
	relation or using a validation set to learn to combine different
	scores,
	we instead learn a small network to predict
	the OKS score
	for each candidate pose $\bar{\mathcal{G}}$
	according to the keypoint heatvalues,
	$h_1(\mathbf{p}_1), h_2(\mathbf{p}_2),
	\dots, h_K(\mathbf{p}_K)$,
	helpful for indicating the visibility,
	as well as the shape feature.
	The shape feature includes
	the distance
	and the relative offset
	between a pair of neighboring keypoints.
	A neighboring pair $(i, j)$
	corresponds to a stick in the COCO dataset,
	and there are $19$ sticks (denoted by
	$\mathcal{E}$) in the COCO dataset.
	The shape feature is denoted as:
	$\{d_{ij} | (i, j) \in \mathcal{E}\}$
	and $\{\mathbf{p}_i - \mathbf{p}_j| (i, j) \in \mathcal{E}\}$.
	The resulting whole feature consists $74$ dimension for the COCO dataset.
	
	We use a small network, 
	consisting two fully-connected layers
	(each followed by a ReLU layer),
	and a linear prediction layer,
	for learning the OKS score
	for a candidate $\bar{\mathcal{G}}$,
	with the real OKS
	as the target.
	We use the pose candidates obtained after 
	grouping over the COCO train2017 dataset
	to form the training examples.
	During inference,
	we feed the shape and heatvalue feature into the small network
	getting the score for each pose candidate.
	This scoring scheme is helpful
	to promote the pose candidates
	that are more  likely  to  be  true  pose.

	\renewcommand{\arraystretch}{1.2}
	\begin{table}[t]
		\centering\setlength{\tabcolsep}{2.6pt}
		\footnotesize
		\caption{GFLOPs and \#parameters 
			of the representative top competitors and our approaches with the backbones: HRNet-W$32$ (H-W$32$), HRNet-W$48$ (H-W$48$) and HrHRNet-W$48$ (Hr-W$48$).
			AE-HG = associative embedding-Hourglass.
		}
		\vspace{-.3cm}
		\label{tab:model-size}
		\begin{tabular}{l|ccc|ccc}
			\hline
			~ & AE-HG & PersonLab & HrHRNet & H-W$32$ & H-W$48$ & Hr-W$48$\\
			\hline
			Input size & $512$ & $1401$ & $640$ & $512$ & $640$ & $640$\\
			\hline
			\#param. (M) & $227.8$ & $68.7$ & $63.8$ & $30.7$ & $66.8$ &$66.9$ \\
			\hline
			GFLOPs & $206.9$ & $405.5$ & $154.3$ & $63.7$ & $170.1$ & $179.5$\\
			\hline 
		\end{tabular} 
	\end{table}

\section{Experiments}
	\label{sec:experiments}
	\subsection{Setting}
	\label{sec:coco}
	\noindent\textbf{Dataset.}
	We evaluate our approach on the COCO keypoint detection task~\cite{LinMBHPRDZ14}. 
	The train$2017$ set includes $57K$ images 
	and $150K$ person instances annotated with $17$ keypoints, 
	the val$2017$ set contains $5K$ images,
	and the test-dev$2017$ set consists of $20K$ images. 
	We train the models on the train$2017$ set and report the results on the val$2017$ and test-dev$2017$ sets.
	
	\vspace{.1cm}
	\noindent\textbf{Evaluation metric.}
	The standard average precision and recall based on Object Keypoint Similarity (OKS) are adopted as the evaluation metrics. 
	Object Keypoint Similarity (OKS):
	$\operatorname{OKS} = \frac{\sum_{i}\exp(-d_i^2/2s^2k_i^2)\delta(v_i > 0)}{\sum_i \delta(v_i > 0)}$, where $d_i$ is the Euclidean distance between each corresponding ground truth and the detected keypoint,
	$v_i$ is the visibility flag of the ground truth,
	$s$ is the object scale, and 
	$k_i$ is a per-keypoint constant that controls falloff.
	We report the following metrics\footnote{\url{http://cocodataset.org/\#keypoints-eval}}:
	$\operatorname{AP}$ 
	(the mean of $\operatorname{AP}$ scores at 
	$\operatorname{OKS} = 0.50, 0.55, \dots,0.90, 0.95$),
	$\operatorname{AP}^{50}$ ($\operatorname{AP}$ at $\operatorname{OKS} = 0.50$),
	$\operatorname{AP}^{75}$ ($\operatorname{AP}$ at $\operatorname{OKS} = 0.75$),
	$\operatorname{AP}^M$  for medium objects,
	$\operatorname{AP}^L$  for large objects,
	and $\operatorname{AR}$ (the mean of $\operatorname{AR}$ scores at $\operatorname{OKS} = 0.50, 0.55, \dots,0.90, 0.95$), 
	$\operatorname{AR}^{50}$ ($\operatorname{AR}$ at $\operatorname{OKS} = 0.50$),
	$\operatorname{AR}^{75}$ ($\operatorname{AR}$ at $\operatorname{OKS} = 0.75$),
	$\operatorname{AR}^{M}$ for medium objects,
	$\operatorname{AR}^{L}$ for large objects.
	
	\vspace{.1cm}
	\noindent\textbf{Training.}
	The data augmentation follows~\cite{NewellHD17}
	and includes
	random rotation ($[\ang{-30}, \ang{30}] $),
	random scale ($[0.75, 1.5]$) and random translation ($[-40, 40]$). 
	We conduct the image cropping
	to $512\times512$ (for HRNet-W$32$) or $640\times640$ (for HRNet-W$48$ and HrHRNet-W$48$) with random flipping as training samples.
	
	We use the Adam optimizer~\cite{KingmaB14}.
	The base learning rate is set as $1\mathrm{e}{-3}$,
	and is dropped to $1\mathrm{e}{-4}$ and $1\mathrm{e}{-5}$ 
	at the $90$th and $120$th epochs, respectively.
	The training process is terminated within $140$ epochs.
	
	\vspace{.1cm}
	\noindent\textbf{Testing.} 
	We resize the short side of the images to $512/640$ and keep the aspect ratio between height and width. Following \cite{NewellHD17}, we adopt three scales $0.5, 1$ and $2$ in multi-scale testing and compute the heatmap and pose positions by averaging the heatmaps and pixel-wise keypoint regressions of the original and flipped images.
	
	\renewcommand{\arraystretch}{1.15}
	\begin{table*}[t]
		\caption{Comparisons on the COCO validation set. $^*$ means using refinement. AE: Associative Embedding~\cite{NewellHD17}.}
		\vspace{-.3cm}
		\centering\setlength{\tabcolsep}{10pt}
		\label{table:coco_val}
		\small
		\begin{tabular}{l|c|lllll|lll}
		    \hline
			Method & Input size &$\operatorname{AP}$ & $\operatorname{AP}^{50}$ & $\operatorname{AP}^{75}$ & $\operatorname{AP}^{M}$ & $\operatorname{AP}^{L}$ & $\operatorname{AR}$ & $\operatorname{AR}^{M}$ & $\operatorname{AR}^{L}$\\
			\hline
			\multicolumn{10}{c}{single-scale testing}\\
			\hline
			CenterNet-DLA~\cite{ZhouWK19} & $512$ &$58.9$ & $-$ & $-$ & $-$ & $-$&$-$ &$-$ & $-$\\
			CenterNet-HG~\cite{ZhouWK19} & $512$ &$64.0$ & $-$ & $-$ & $-$ & $-$ & $-$ & $-$& $-$\\
			PifPaf~\cite{KreissBA19} & $-$   
			&$67.4$ & $-$ & $-$ & $-$ & $-$ & $-$ & $-$& $-$\\
			PersonLab~\cite{PapandreouZCGTM18} & 601 &$54.1$ & $76.4$ & $57.7$ & $40.6$ & $73.3$ & $57.7$ & $43.5$& $77.4$\\
			PersonLab~\cite{PapandreouZCGTM18} & 1401 &$66.5$ & $86.2$ & $71.9$ & $62.3$ & $73.2$ & $70.7$ & $65.6$ & $77.9$\\
			HrHRNet-W$32$ + AE ~\cite{cheng2019bottom} & $512$ & $67.1$ & $86.2$ & $73.0$ & $-$ & $-$ & $-$ & $61.5$ & $76.1$ \\
			HrHRNet-W$48$ + AE ~\cite{cheng2019bottom} & $640$ & $69.9$ & $87.2$ & $76.1$ & $-$ & $-$ & $-$ & $65.4$ & $76.4$ \\
			\hline 
			Ours (HRNet-W$32$) & $512$ & $67.8$ & $86.8$ & $74.0$ & $62.0$ & $76.4$ & $72.3$ & $65.6$ & $82.0$\\
			Ours (HRNet-W$48$) & $640$ & $70.1$ & $88.1$ & $76.0$ & $65.6$ & $77.2$ & $74.8$ & $69.2$ & $82.9$\\
			Ours (HrHRNet-W$48$) & $640$ & $71.3$& $88.4$ & $77.0$ & $67.5$ & $77.3$ & $75.8$ & $70.9$ & $83.1$\\
			\hline
			\multicolumn{10}{c}{multi-scale testing}\\
			\hline
			Deep body-foot~\cite{Hidalgo_2019_ICCV} & $480$ & $66.4$ & $-$ & $-$ & $-$ & $-$ & $-$ & $-$ & $-$ \\
			HrHRNet-W$32$ + AE ~\cite{cheng2019bottom} & $512$ & $69.9$ & $87.1$ & $76.0$ & $-$ & $-$ & $-$ & $65.3$ & $77.0$ \\
			HrHRNet-W$48$ + AE ~\cite{cheng2019bottom} & $640$ & $72.1$ & $88.4$ & $78.2$ & $-$ & $-$ & $-$ & $67.8$ & $78.3$ \\
			\hline
			Ours (HRNet-W$32$) & $512$ & $70.7$ & $88.0$ & $76.9$ & $66.1$ & $77.7$ & $75.8$ & $70.2$ & $83.8$\\
			Ours (HRNet-W$48$) & $640$ & $72.5$ & $88.9$ & $78.7$ & $68.9$ & $78.2$ & $77.7$ & $72.8$ & $84.7$\\
			Ours (HrHRNet-W$48$) & $640$ & $72.9$ & $89.2$ & $78.8$ & $69.3$ & $78.5$ & $78.2$ & $73.2$ & $85.4$\\
			\hline
		\end{tabular}
	\end{table*}
	
	\begin{table*}[t]
		\caption{Comparisons on the COCO test-dev set. $^*$ means using refinement. AE: Associative Embedding.}
		\vspace{-.3cm}
		\centering\setlength{\tabcolsep}{10pt}
		\label{table:coco_test}
		\small
		\begin{tabular}{l|c|lllll|lll}
			\hline
			Method & Input size &$\operatorname{AP}$ & $\operatorname{AP}^{50}$ & $\operatorname{AP}^{75}$ & $\operatorname{AP}^{M}$ & $\operatorname{AP}^{L}$ & $\operatorname{AR}$ & $\operatorname{AR}^{M}$ & $\operatorname{AR}^{L}$\\
			\hline
			\multicolumn{10}{c}{single-scale testing}\\
			\hline
			OpenPose$^*$~\cite{CaoSWS17} & $-$ &$61.8$ & $84.9$&$67.5$&$57.1$&$68.2$&$66.5$ & $-$& $-$\\
			AE~\cite{NewellHD17} & $512$ &$56.6$ & $81.8$ & $61.8$ & $49.8$ & $67.0$ & $-$ & $-$&$-$ \\
			CenterNet-DLA~\cite{ZhouWK19} & $512$ &$57.9$ & $84.7$ & $63.1$ & $52.5$ & $67.4$ & $-$ & $-$& $-$\\
			CenterNet-HG~\cite{ZhouWK19} & $512$ &$63.0$ & $86.8$ & $69.6$ & $58.9$ & $70.4$ & $-$ & $-$& $-$\\
			PifPaf~\cite{KreissBA19} & $-$   
			&$66.7$ & $-$ & $-$ & $62.4$ & $72.9$ & $-$& $-$& $-$\\
			SPM$^*$~\cite{NieZYF19} & - & $66.9$ & $88.5$ & $72.9$ & $62.6$ & $73.1$ & $-$ & $-$& $-$\\
			PersonLab~\cite{PapandreouZCGTM18} & $1401$ &$66.5$ & $88.0$ & $72.6$ & $62.4$ & $72.3$ & $71.0$ & $66.1$&$77.7$\\
			HrHRNet-W$48$ + AE ~\cite{cheng2019bottom} & $640$ & $68.4$ & $88.2$ & $75.1$ & $64.4$ & $74.2$ & $-$ & $-$ & $-$ \\
			\hline 
			Ours (HRNet-W$32$) & $512$ & $66.6$ & $87.8$ & $72.8$ & $61.1$ & $74.5$ & $71.4$ & $64.6$ & $80.8$\\
			Ours (HRNet-W$48$) & $640$ & $69.4$ & $88.9$ & $76.2$ & $64.9$ & $75.7$ & $74.3$ & $68.5$ & $82.2$\\
			Ours (HrHRNet-W$48$) & $640$ & $70.2$ & $89.5$ & $77.3$ & $66.5$ & $75.6$ & $75.1$ & $70.1$ & $82.1$\\
			\hline
			\multicolumn{10}{c}{multi-scale testing}\\
			\hline
			AE~\cite{NewellHD17} & $512$ &$63.0$ & $85.7$ & $68.9$ & $58.0$ & $70.4$ & $-$ & $-$& $-$ \\
			AE$^*$~\cite{NewellHD17} & $512$ &$65.5$ & $86.8$&$72.3$&$60.6$&$72.6$&$70.2$ & $64.6$& $78.1$\\
			PersonLab~\cite{PapandreouZCGTM18} & $1401$ &$68.7$ & $89.0$&$75.4$&$64.1$&$75.5$&$75.4$ & $69.7$& $83.0$\\
			HrHRNet-W$48$ + AE~\cite{cheng2019bottom} & $640$ & $70.5$ & $89.3$ & $77.2$ & $66.6$ & $75.8$ & $-$& $-$& $-$ \\
			\hline
			Ours (HRNet-W$32$) & $512$ & $69.4$ & $88.9$ & $76.2$ & $64.9$ & $75.8$ & $74.9$ & $69.1$ & $82.9$\\
			Ours (HRNet-W$48$) & $640$ & $71.4$ & $89.8$ & $78.3$ & $67.8$ & $76.8$ & $76.9$ & $71.7$ & $84.1$\\
			Ours (HrHRNet-W$48$) & $640$ & $71.8$ & $90.2$ & $78.7$ & $68.3$ & $76.8$ & $77.4$ & $72.4$ & $84.3$\\
			\hline
		\end{tabular}
	\end{table*}

	\begin{table}[t]
		\centering
        \setlength{\tabcolsep}{3.3pt}
        \scriptsize
		{
            \renewcommand{\arraystretch}{1.3}
			\newcommand{\tabincell}[2]{\begin{tabular}{@{}#1@{}}#2\end{tabular}}
			\caption{Ablation study: heatmap-guided pixel-wise keypoint regression, heatmap tradeoff loss,
				adaptive representation transformation (ART) and scoring. 
				Scoring only affects the overall quality.}
				\vspace{-.3cm}
			\begin{tabular}{ c | c | c | c | c | c | c  }
			\hline
				\pbox[c]{20cm}{Heatmap \\ guidance} & 
				\pbox[c]{20cm}{Tradeoff \\ loss} &
				\pbox[c]{20cm}{Adaptive \\ representation} &
				\pbox[c]{20cm}{Scoring\\} &
				\pbox[c]{20cm}{Regression \\ quality} &
				\pbox[c]{20cm}{Heatmap \\ quality} &
				\pbox[c]{20cm}{Final \\ quality} \\
				\hline
				& & & & $59.6$ & $71.3$ & $64.5$ \\
				\checkmark & & & & $60.5$ & $71.8$ & $64.9$ \\ 
				\checkmark & \checkmark & & & $61.5$ & $73.2$ & $66.2$ \\
				\checkmark & \checkmark & \checkmark & & $65.1$ & $73.3$ &  $67.2$ \\
				\checkmark & \checkmark & \checkmark & \checkmark &  $65.1$ & $73.3$ & $67.8$\\
				\hline
			\end{tabular}
			\label{tab:ablation-study}
		}
		\vspace{-.2cm}
	\end{table}

	\subsection{Results}
	
	\noindent\textbf{Validation results.} Table \ref{table:coco_val} 
	shows the comparisons of our method 
	and other state-of-the-art methods.
	We use HRNet-W$32$ and HRNet-W$48$
	as the backbones and adopt three parallel branches each estimating the keypoint heatmap. We average the three output heatmaps as the final heatmap prediction
	that, as a part of input, is fed into the pixel-wise keypoint regressor.
	We also test the performance
	using HrHRNet-W$48$ as the backbone\footnote{
		We perform keypoint heatmap estimation
		and pixel-wise keypoint regression 
		over the $4\times$ representation
		for generating the grouping cues.
		We use the $2\times$ resolution representation in HrHRNet-W$48$
		to estimate $2\times$ resolution heatmaps,
		and find the keypoint candidates from the $2\times$ resolution heatmaps
		for the further grouping.}
	Table \ref{tab:model-size} presents
	the parameter
	and computation complexities 
	for our approach and the representative top competitors,
	AE-Hourglass~\cite{NewellHD17}, PersonLab~\cite{PapandreouZCGTM18} and HrHRNet~\cite{cheng2019bottom}.
	
	Our approach,
	using HRNet-W$32$ as the backbone, 
	achieves $67.8$ AP score. 
	Compared to the methods with similar GFLOPs, 
	CenterNet-DLA~\cite{ZhouWK19} and PersonLab~\cite{PapandreouZCGTM18} (with the input size $601$), 
	our approach achieves over $8.9$ improvement.
	In comparison to 
	CenterNet-HG~\cite{ZhouWK19} whose model size is far larger than HRNet-W$32$,
	our gain is $3.8$,
	consisting of two aspects:
	(1) $0.6$\footnote{
		In the gain $0.6$,
		$0.3$ comes from using the average of keypoint positions as the center position 
		compared to using the human box center~\cite{ZhouWK19}.} from our baseline
	(our baseline using HRNet-W$32$ as the backbone
	achieves $64.6$ (Table~\ref{tab:ablation-study})
	(2) the remaining gain $3.2$ from our methodology (Table~\ref{tab:ablation-study}).

	Our approach benefits from large input size, large model size and higher resolution representations. 
	Our approach, with HRNet-W$48$ as the backbone,
	the input size $640$, 
	obtains the best performance $70.1$
	and $2.3$ gain over HRNet-W$32$. 
	Compared with state-of-the-art methods, 
	our approach gets $6.1$
	gain over CenterNet-HG, $3.6$ gain over PersonLab (the input size $1401$), 
	and $2.7$ gain over PifPaf~\cite{KreissBA19} 
	whose GFLOPs are more than twice as many as ours.
	Besides,
	we use the higher resolution representation (HrHRNet-W$48$~\cite{cheng2019bottom}),
	leading to $1.2$ gain
	over HRNet-W$48$.
	
	Following~\cite{NewellHD17,PapandreouZCGTM18}, we report
	the results with multi-scale testing. 
	This brings about $2.9$ gain for HRNet-W$32$, 
	$2.4$ gain for HRNet-W$48$ and $1.6$ points for HrHRNet-W$48$.
	
	\vspace{.1cm}
	\noindent\textbf{Test-dev results.} 
	The results of our approach and other state-of-the-art methods 
	on the test-dev dataset
	are presented in Table \ref{table:coco_test}. 
	
	Our approach 
	with HRNet-W$32$ as the backbone
	achieves $66.6$ AP scores, 
	and significantly outperforms the methods with the similar model size. Our approach with HrHRNet-W$48$ as the backbone 
	gets the best performance $70.2$, 
	leading to $3.7$ gain over PersonLab, $3.5$ gain over PifPaf~\cite{KreissBA19}, 
	and $1.8$ gain over HrHRNet~\cite{cheng2019bottom}. 
	With multi-scale testing, 
	our approach
	with HRNet-W$32$ as the backbone
	achieves $69.4$, 
	even better than PersonLab with a much larger model size. 
	Our approch with HrHRNet-W$48$ achieves $71.8$ AP score, 
	much better than associative embedding~\cite{NewellHD17}, $3.1$ gain over PersonLab, and $1.3$ gain over HrHRNet~\cite{cheng2019bottom}.
	
	\begin{figure*}[ht!]
		\small
		\centering
		\includegraphics[height = 0.135\textwidth]{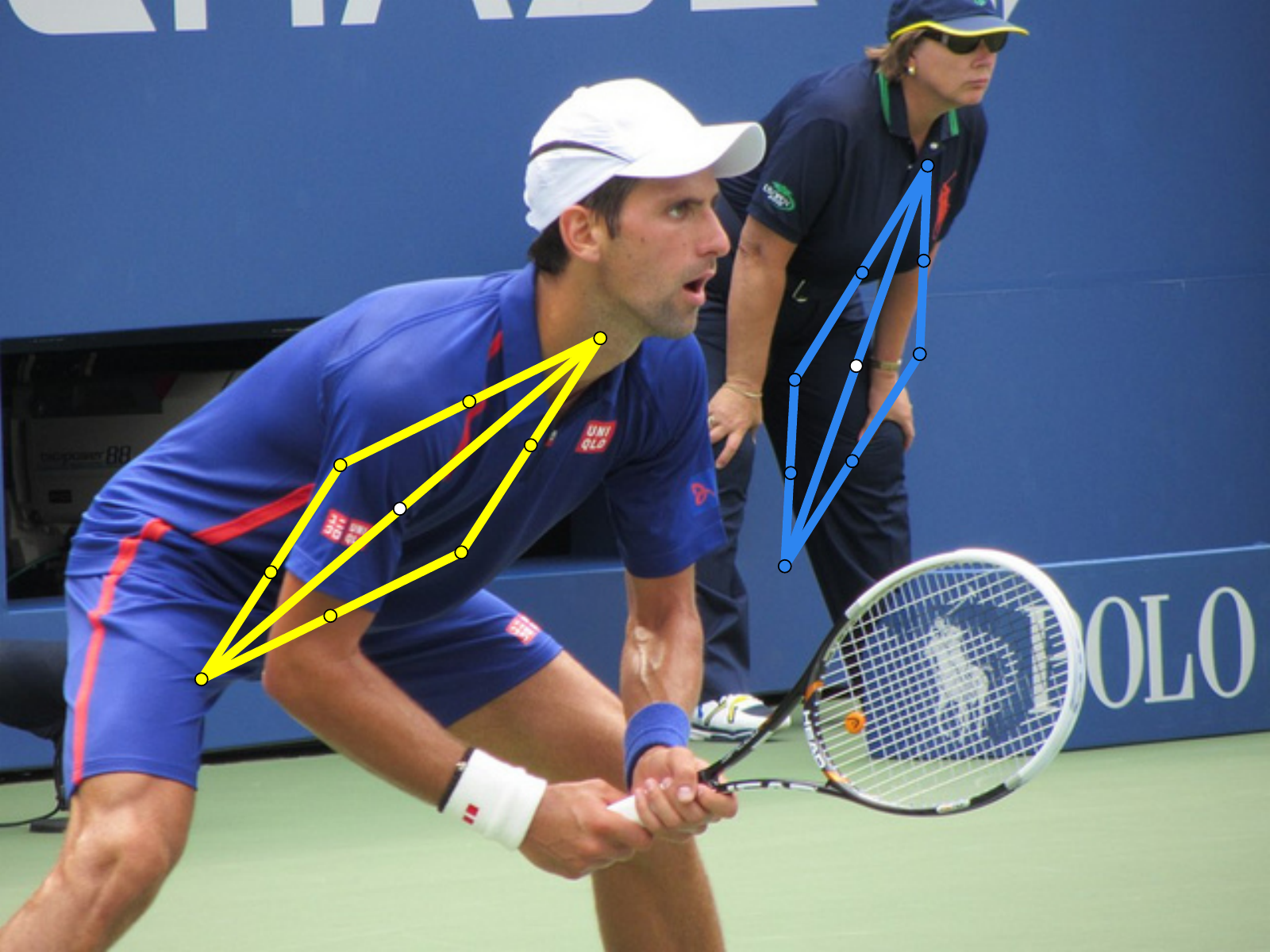}
		\includegraphics[height = 0.135\textwidth]{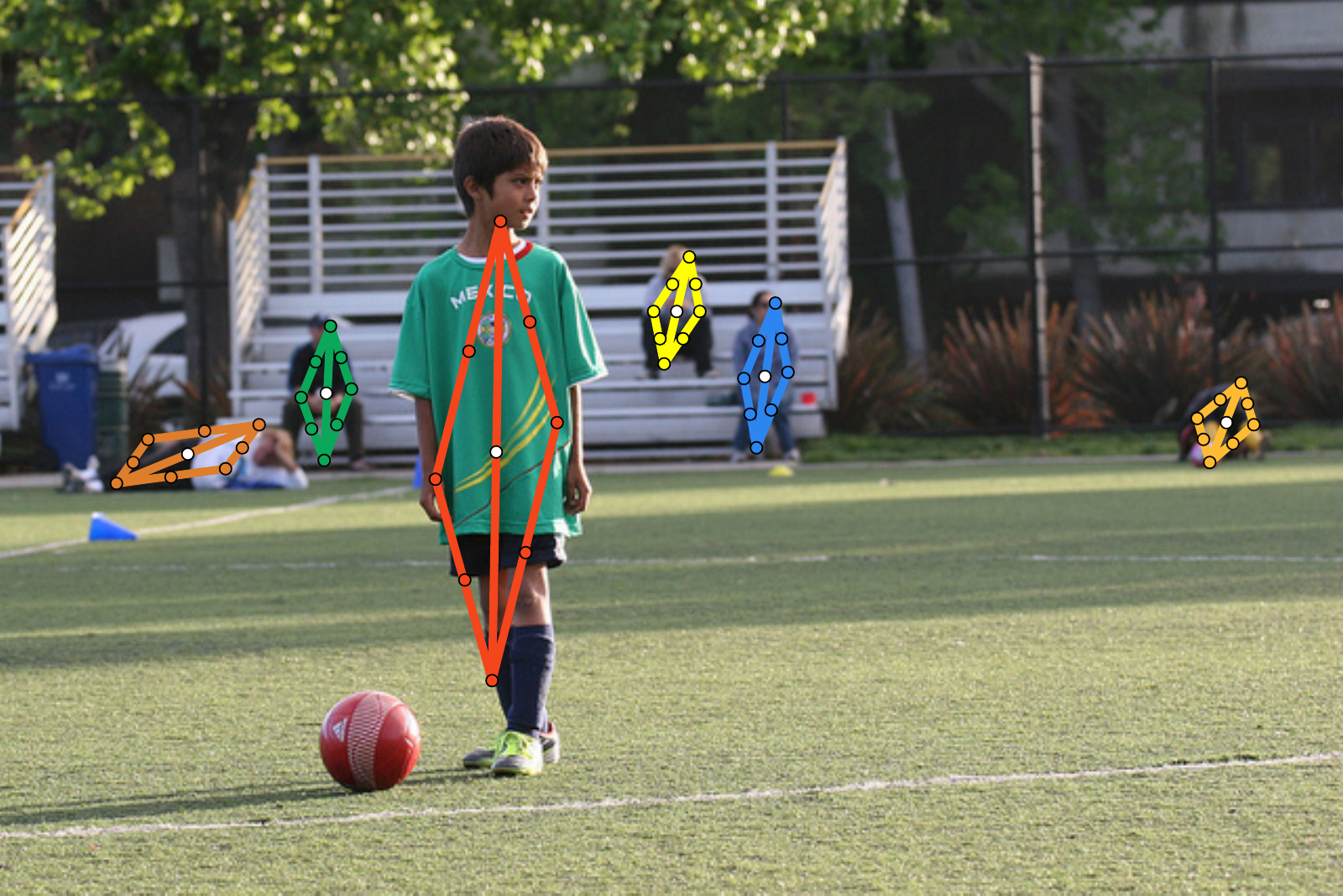}
		\includegraphics[height = 0.135\textwidth]{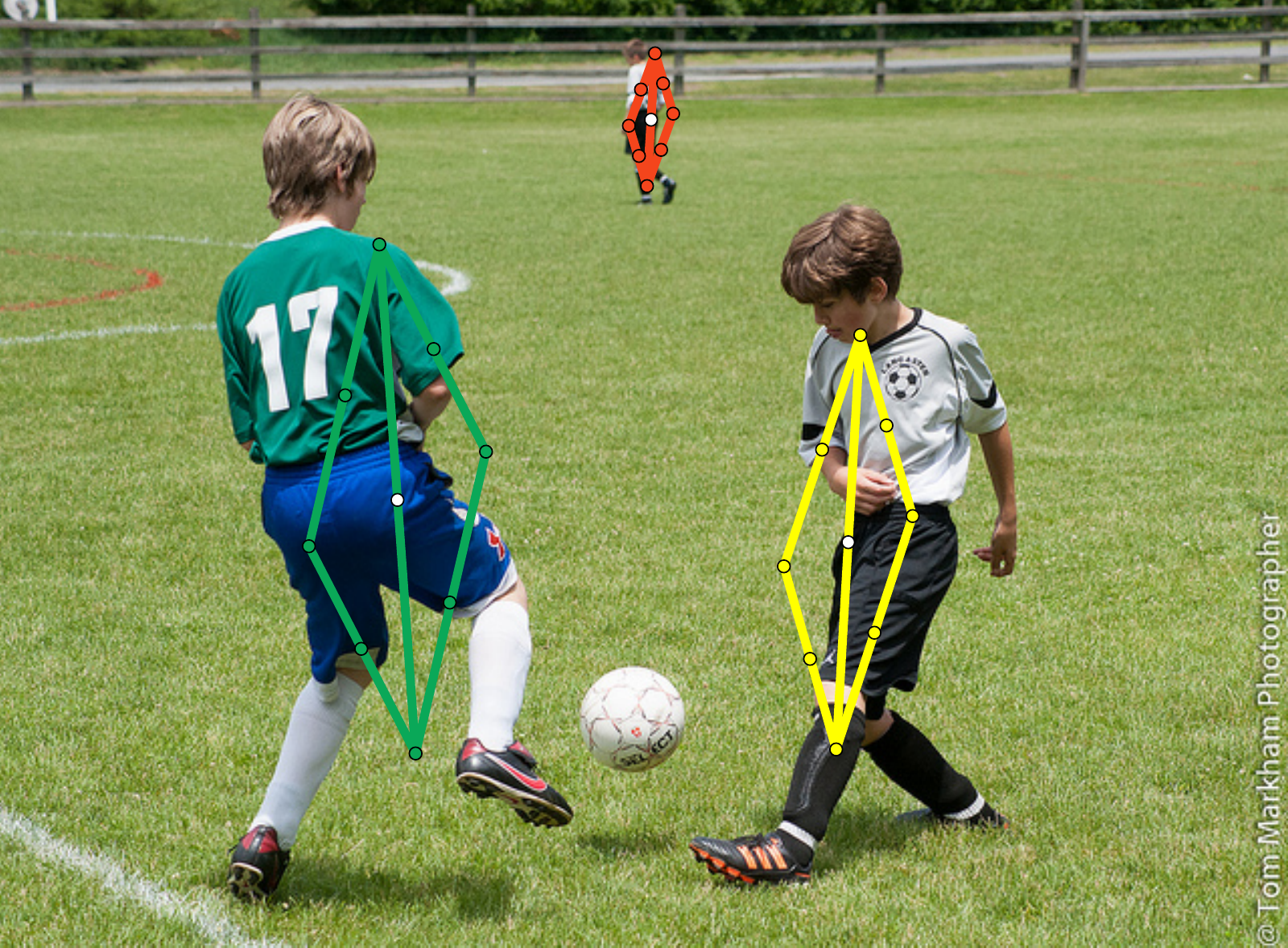}
		\includegraphics[height = 0.135\textwidth]{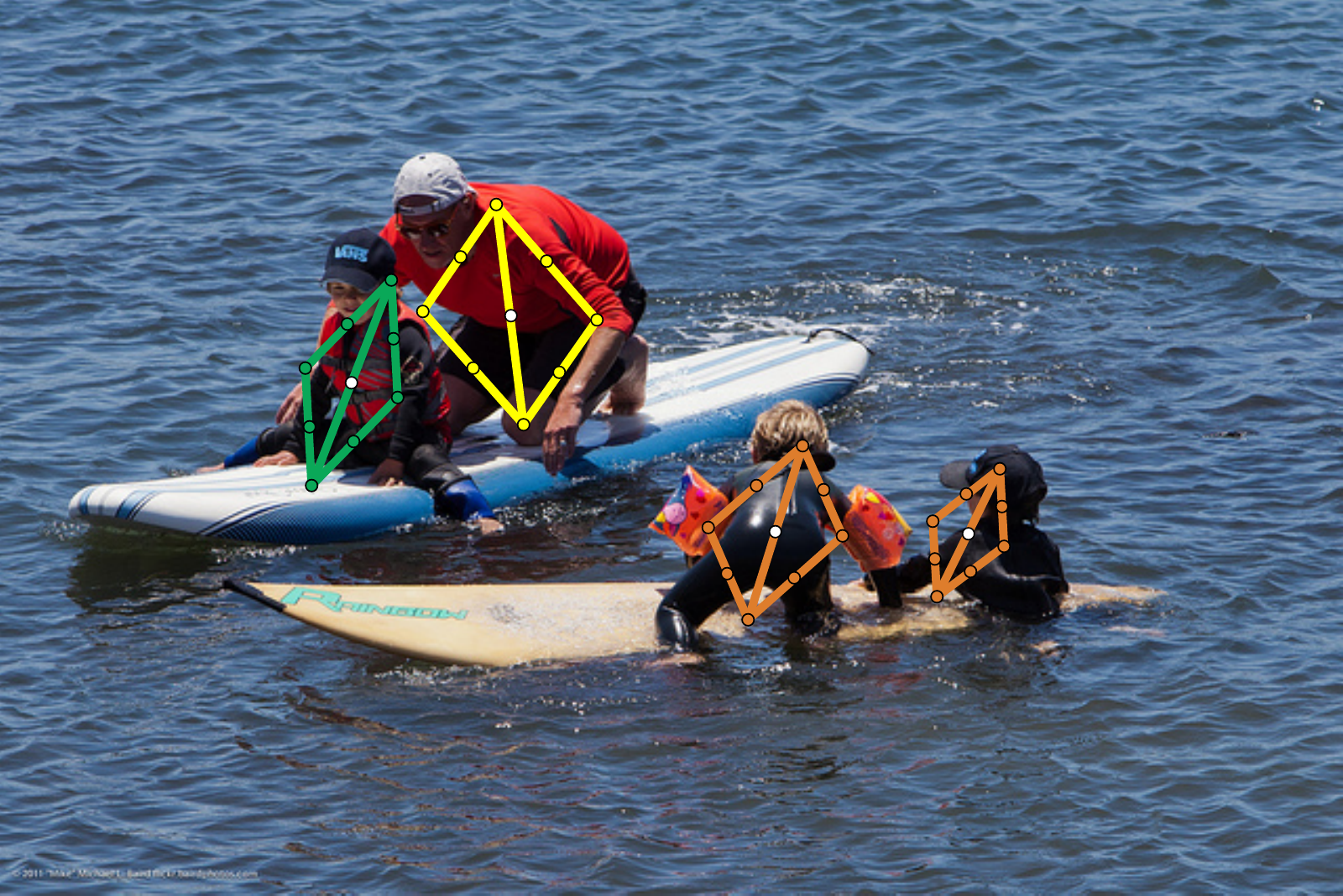}
		\includegraphics[height = 0.135\textwidth]{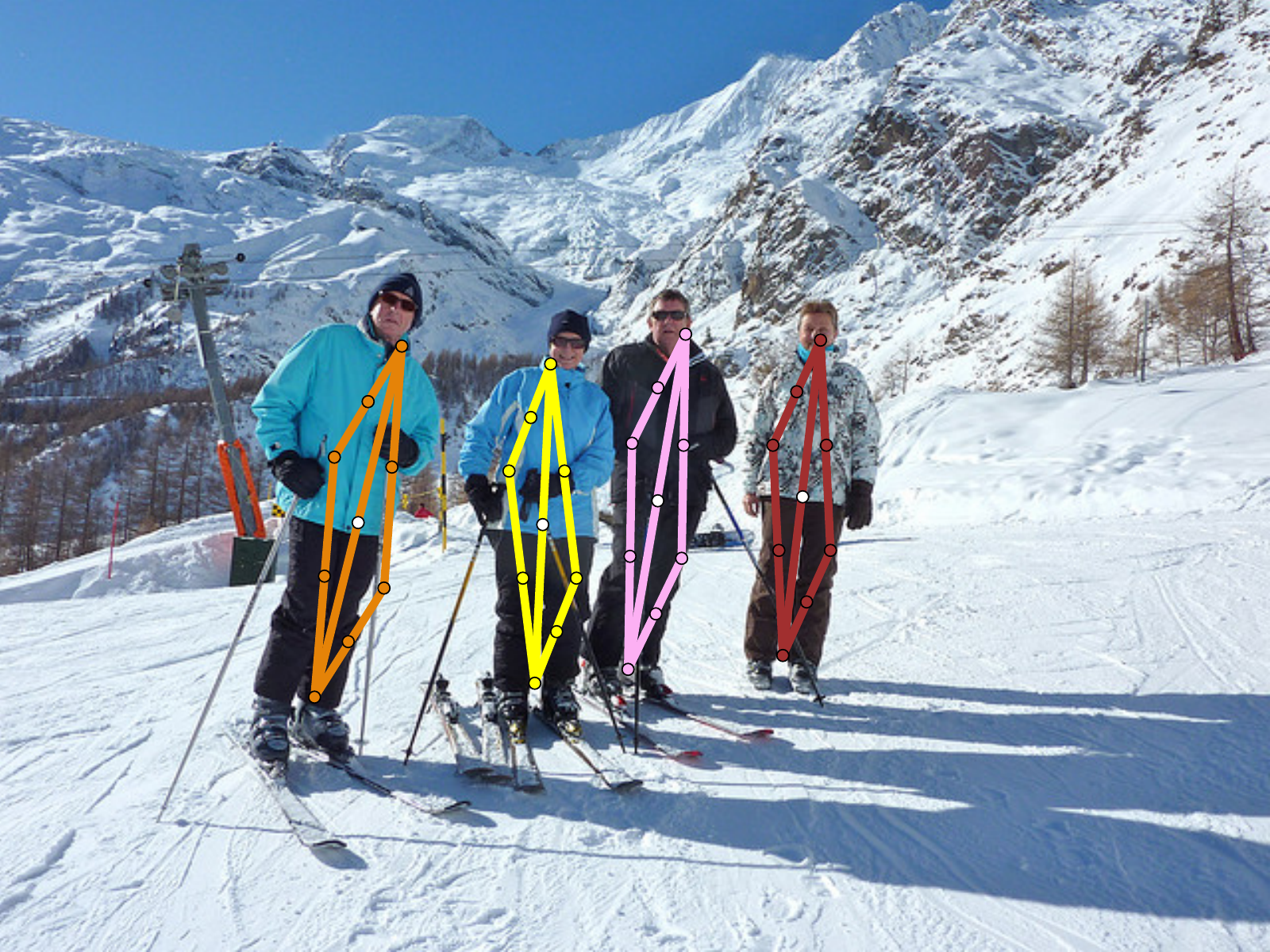}

\vspace{-.3cm}
		\caption{Illustrating the pixel-wise STN contained in the pixel-wise keypoint regression head. 
			We show the $9$ positions learned from pixel-wise STN at the pose center. 
		}
		\label{fig:illustrationSTN}
	\end{figure*}
	
	\begin{figure*}[t]
		\small
		\centering
		
		\subfloat[b]{\includegraphics[width=1.57in]{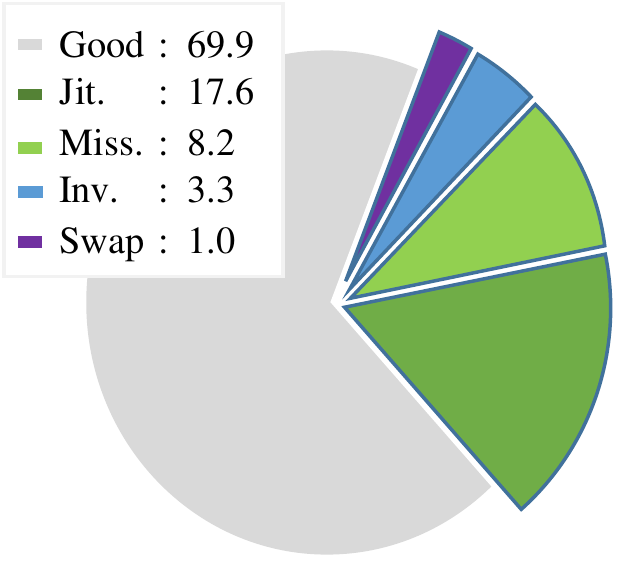}}~~~~~~
		\subfloat[b+c1]{\includegraphics[width=1.57in]{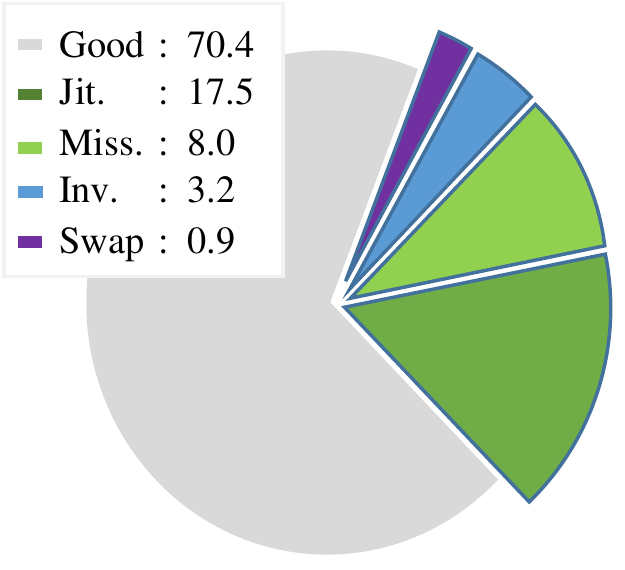}}~~~~~~
		\subfloat[b+c1+c2]{\includegraphics[width=1.57in]{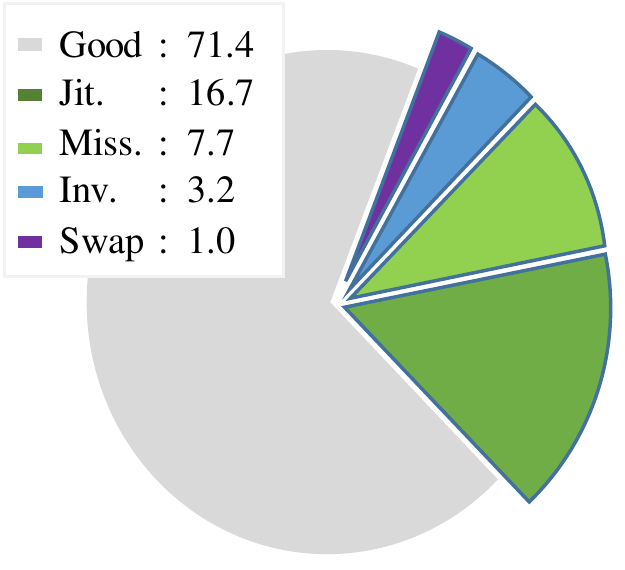}}~~~~~~
		\subfloat[b+c1+c2+c3]{\includegraphics[width=1.57in]{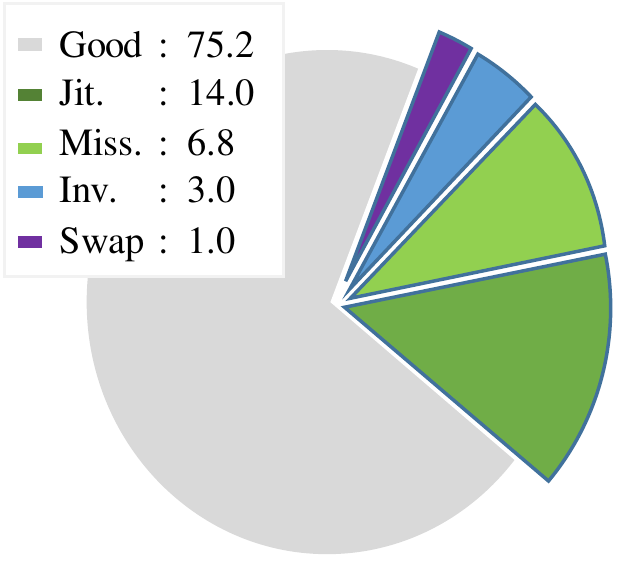}}
		
		\subfloat[b]{\includegraphics[width=1.57in]{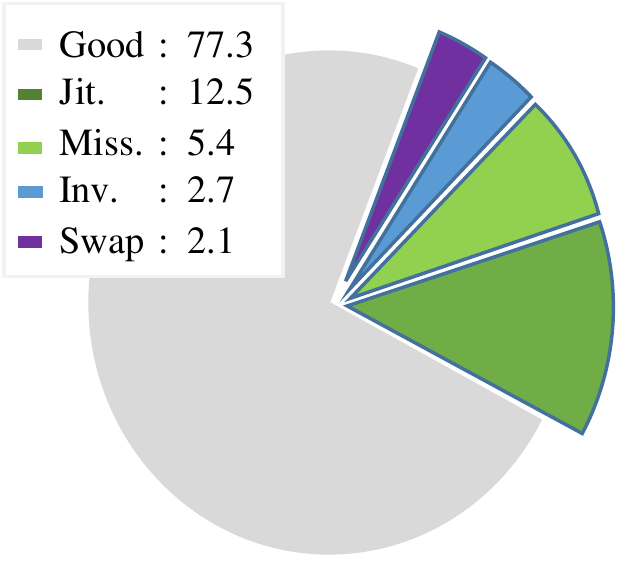}}~~~~~~
		\subfloat[b+c1]{\includegraphics[width=1.57in]{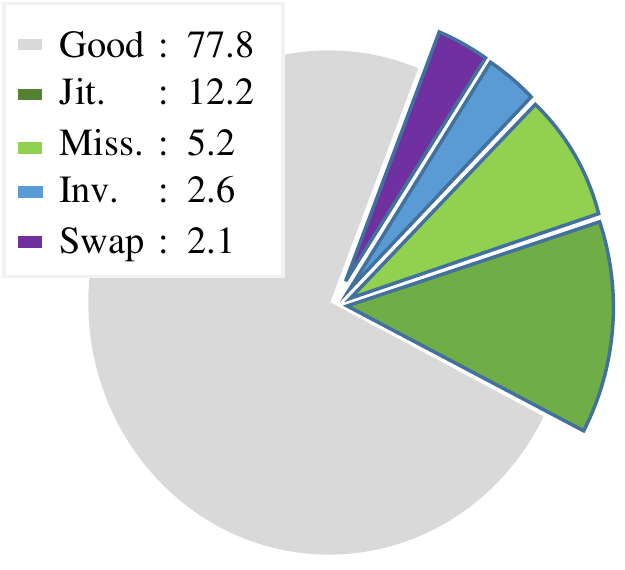}}~~~~~~
		\subfloat[b+c1+c2]{\includegraphics[width=1.57in]{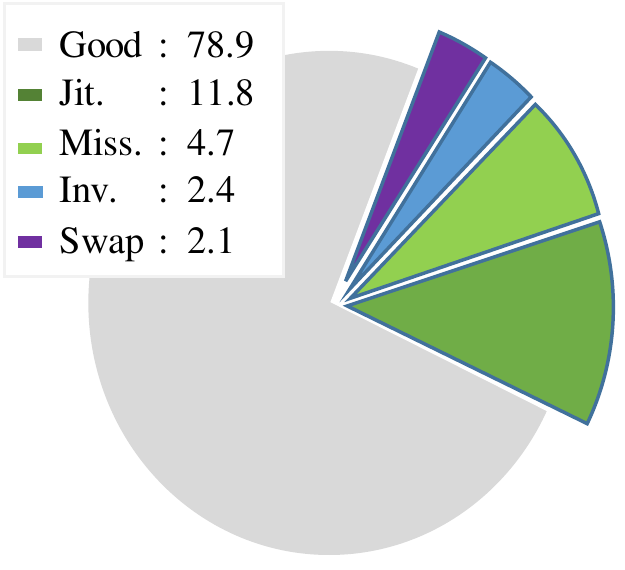}}~~~~~~
		\subfloat[b+c1+c2+c3]{\includegraphics[width=1.57in]{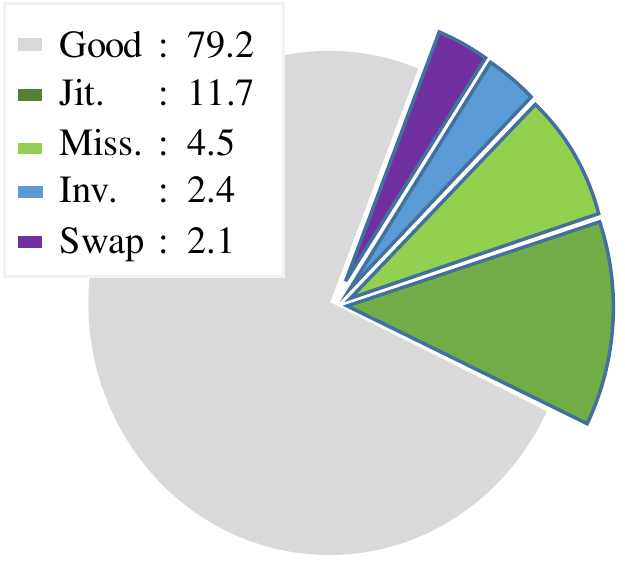}}
		
		\subfloat[b]{\includegraphics[width=1.57in]{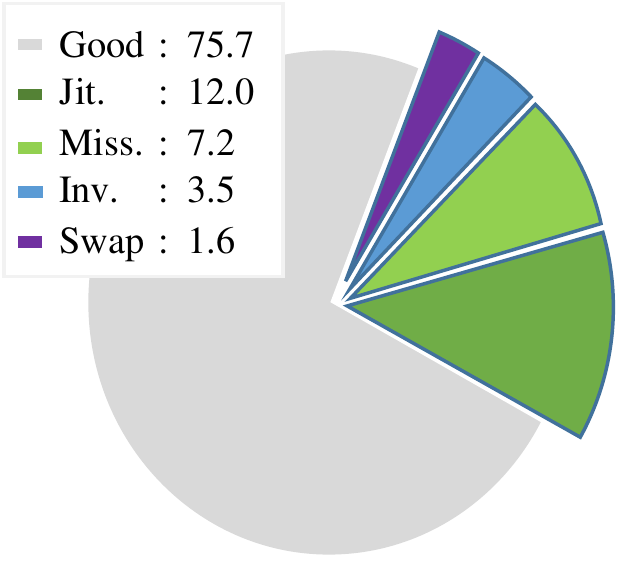}}~~~~~~
		\subfloat[b+c1]{\includegraphics[width=1.57in]{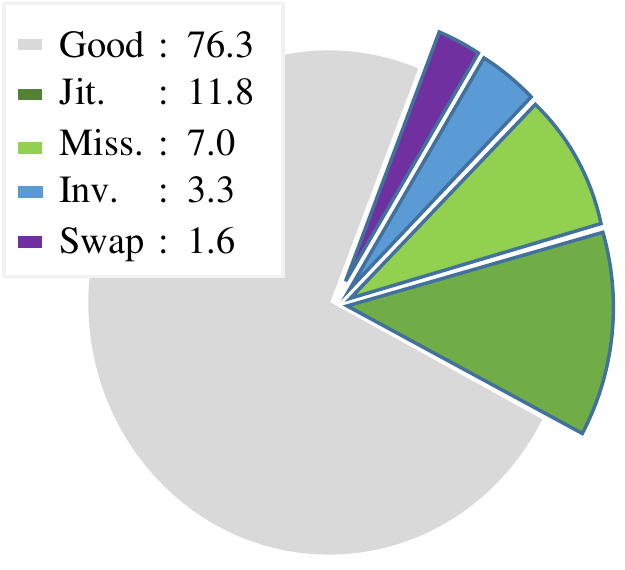}}~~~~~~
		\subfloat[b+c1+c2]{\includegraphics[width=1.57in]{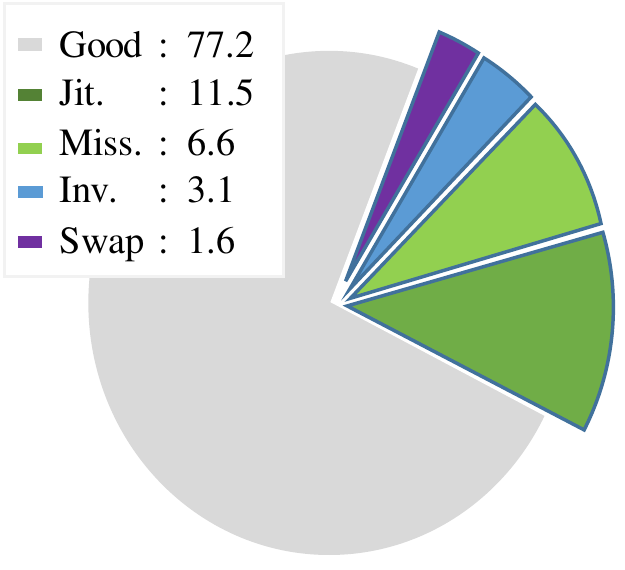}}~~~~~~
		\subfloat[b+c1+c2+c3]{\includegraphics[width=1.57in]{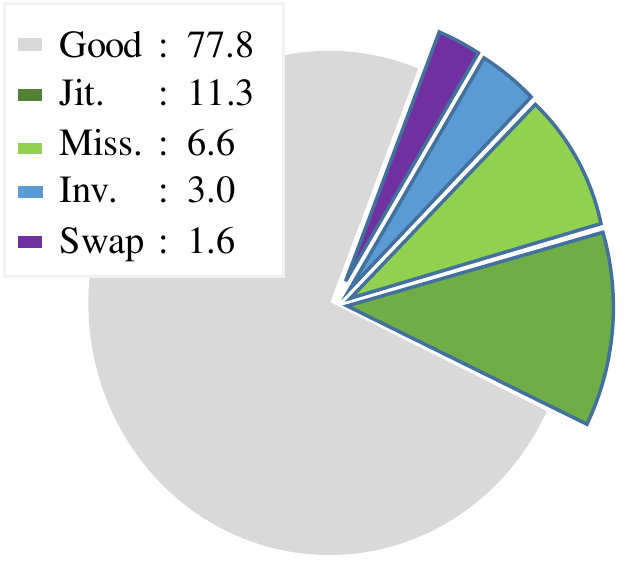}}
		
		\vspace{-.3cm}
		\caption{Component analysis in terms of
			four errors
			for three qualities:
			(a) - (d): pixel-wise keypoint regression;
			(e) - (h): keypoint heatmap estimation;
			(i) - (l): final poses.
			b: the baseline corresponding to the $64.6$ AP score
			in Table \ref{tab:ablation-study}; 
			c1: heatmap guided pixel-wise keypoint estimation;
			c2:  heatmap tradeoff loss; 
			c3: adaptive representation transformation.}
		\label{fig:erroranalysis}
	\end{figure*}

\vspace{-.1cm}
\subsection{Ablation Study}

	We study the effects of the components in our approach:
	heatmap-guided pixel-wise keypoint regression,
	heatmap tradeoff loss,
	adaptive representation transformation,
	and pose scoring.
	We check three qualities:
	pixel-wise keypoint regression quality,
	heatmap estimation quality,
	and final quality.
	The first one
	is obtained
	by directly using the regression results
	and evaluating it using the AP scores.
	The second one is done by 
	grouping keypoints identified from heatmaps
	using the groundtruth poses as the grouping cue
	(i.e., replacing the regressed poses).
	The final quality is the quality of the whole scheme
	of our approach.
	
	The ablation study result is presented in Table~\ref{tab:ablation-study}.
	The heatmap guidance scheme indeed boosts
	pixel-wise keypoint regression:
	the improvements is $0.9$.
	The final quality is improved by $0.4$.
	The heatmap tradeoff loss improves the heatmap
	quality greatly (by $1.4$).
	The adaptive representation transformation improves the pixel-wise keypoint regression quality (by $3.6$), and the final results (by $1.0$).
	The scoring scheme ranks the final estimations
	and achieves a gain $0.6$.

	Figure~\ref{fig:erroranalysis}
	illustrates the error analysis results
	for pixel-wise keypoint regression ({\color{red}a - d}),
	keypoint heatmap estimation ({\color{red}e - h})
	and final predictions ({\color{red}i - l}).
	The detailed analysis
	is given 
	in the following.

\vspace{.1cm}
\noindent\textbf{Error analysis.}
	We analyze how each component contributes
	to the performance improvement
	by using the coco analysis tool~\cite{RonchiP17}. Four error types are studied: (i) \textit{Jitter} error: small localization error around the correct keypoint location; (ii) \textit{Miss} error: large localization error, the detected keypoint is not within the proximity of any ground truth keypoint of any instance; (iii) \textit{Inversion} error: confusion between keypoints within a instance. The detected keypoint is in the proximity of a wrong ground truth keypoint belonging to the same instance. (iv) \textit{Swap} error: confusion between keypoints of different instances. The detected keypoint is in the proximity of a ground truth keypoint belonging to a different instance. 
	The detailed definitions are in~\cite{RonchiP17}.
	
\vspace{.1cm}
\noindent\emph{Heatmap guidance:}
	The comparison between 
	Figure \ref{fig:erroranalysis} ({\color{red}a})
	and Figure \ref{fig:erroranalysis} ({\color{red}b})
	shows that
	exploiting the heatmap for pixel-wise keypoint regression
	brings about improvement for all the errors.
	The comparison between 
	Figure \ref{fig:erroranalysis} ({\color{red}e})
	and Figure \ref{fig:erroranalysis} ({\color{red}f}),
	shows the heatmap estimation quality improvement,
	indicating that 
	exploiting the heatmap for improving pixel-wise keypoint regression
	in turns benefits the heatmap estimation,
	though the improvement is not
	as great as pixel-wise keypoint regression.
	
\vspace{.1cm}
\noindent\emph{Heatmap tradeoff loss:}
	The tradeoff aims to balance
	the numbers of 
	the keypoint pixels
	and the non-keypoint pixels,
	mainly for keypoint heatmap estimation.
	This is able to strengthen 
	the keypoint classification capability.
	By comparing 
	Figure \ref{fig:erroranalysis} ({\color{red}g})
	and Figure \ref{fig:erroranalysis} ({\color{red}f}),
	we can see that 
	the missing error is reduced the greatest (by $0.5$).
	This is as expected,
	and the error reduction mainly comes from the classification capability.
	Because of the assistance from the high-quality heatmaps,
	the pixel-wise keypoint regression quality is also 
	improved as seen in Figure \ref{fig:erroranalysis} ({\color{red}c}).
	
\vspace{.1cm}
\noindent\emph{Adaptive representation transformation:}
	The comparison between
	Figure \ref{fig:erroranalysis} ({\color{red}d})
	and Figure \ref{fig:erroranalysis} ({\color{red}c})
	shows that the pixel-wise keypoint regression quality
	is largely improved,
	mainly on the jitter error and the missing error.
	Figure~\ref{fig:illustrationSTN} illustrates
	the sampled $9$ positions at the pose center
	computed by the pixel-wise STN for the pixel-wise keypoint regression head,
	which shows that the human scale and the human rotation
	are captured by the adaption scheme.
	The adaption scheme helps improve the pixel-wise representation by seeing the global context 
	instead of 
	the local context in a regular convolution.

\begin{figure*}[ht!]
	\small
	\centering
	\includegraphics[width=0.24\textwidth,height = 0.183\textwidth]{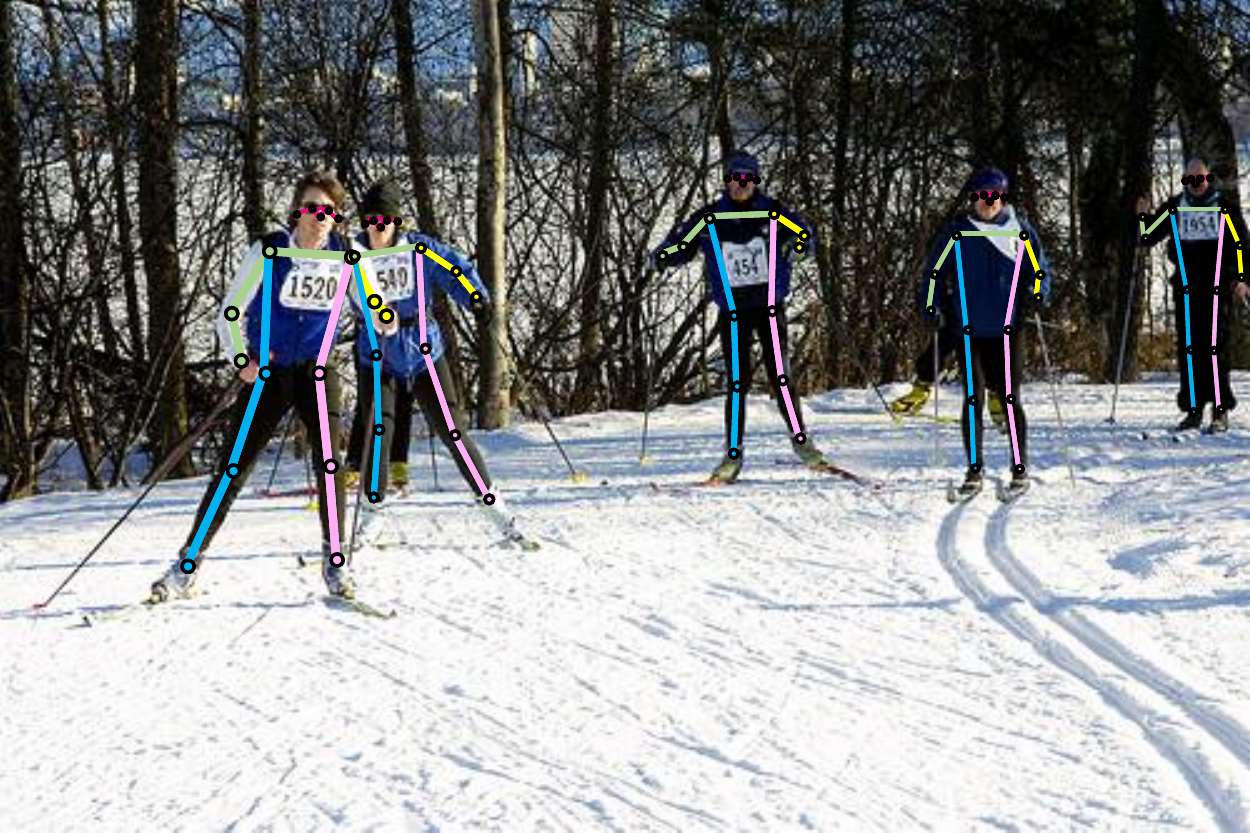}
	\includegraphics[width=0.24\textwidth,height = 0.183\textwidth]{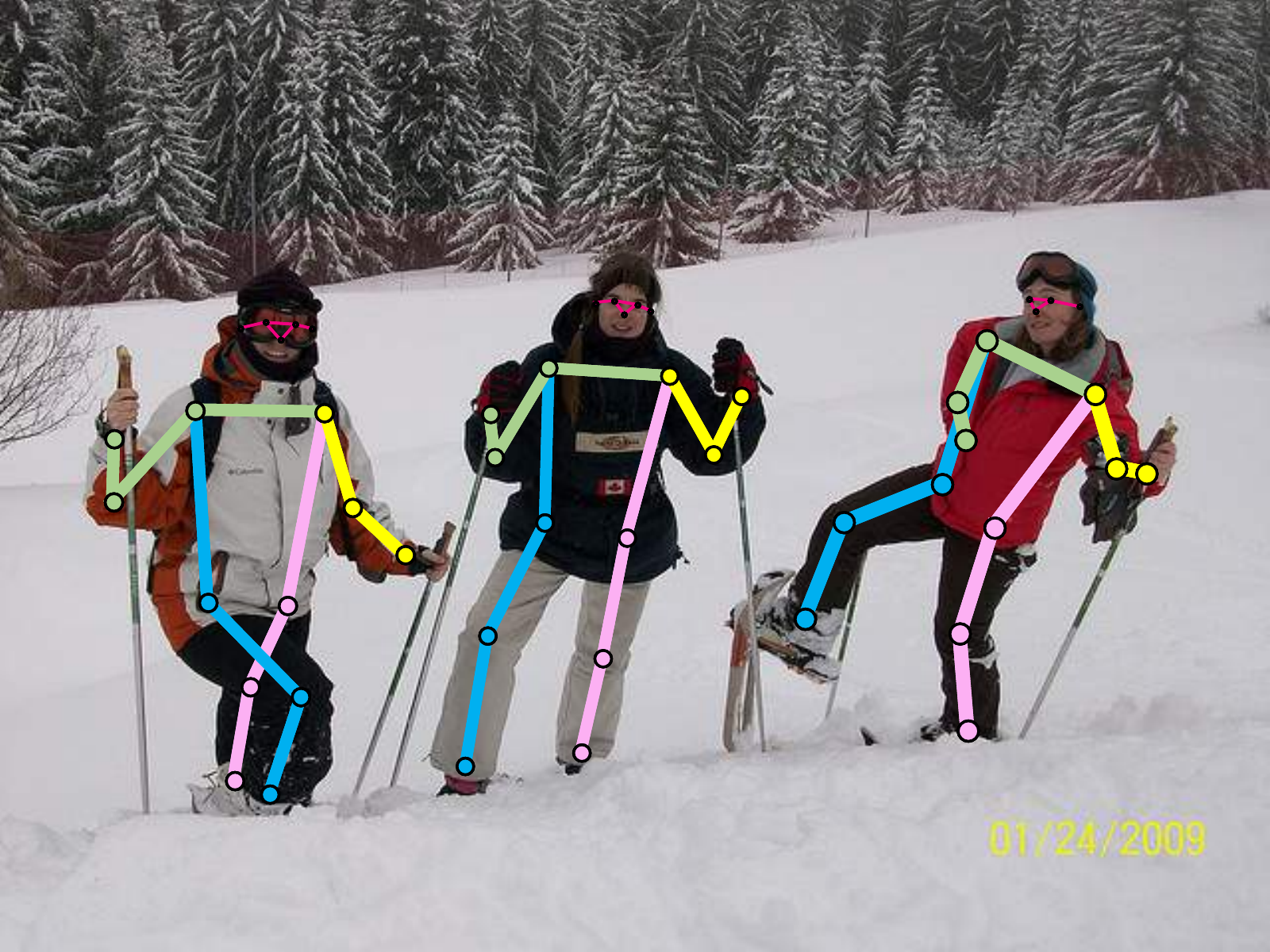}
	\includegraphics[width=0.24\textwidth,height = 0.183\textwidth]{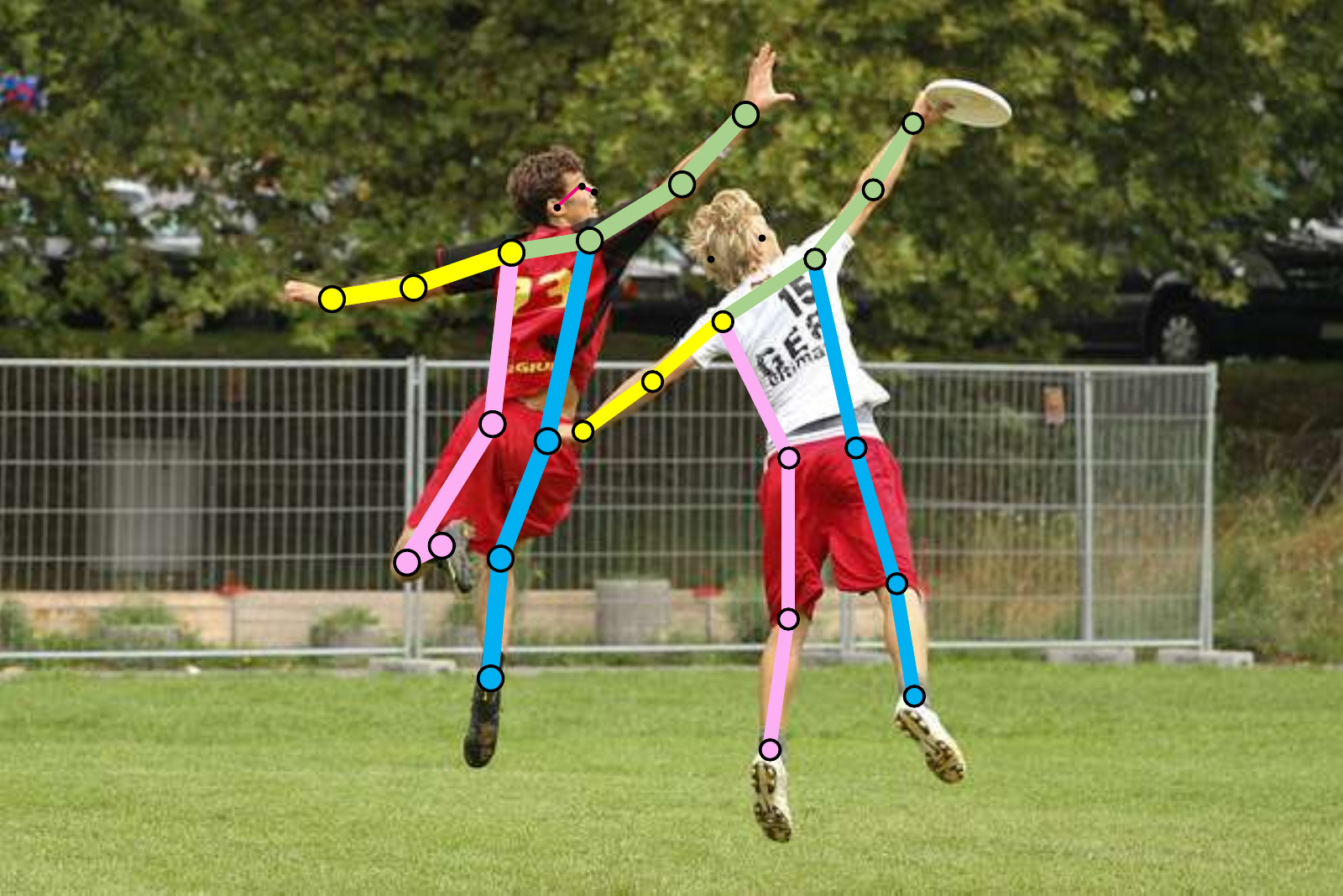}
 	\includegraphics[width=0.24\textwidth,height = 0.183\textwidth]{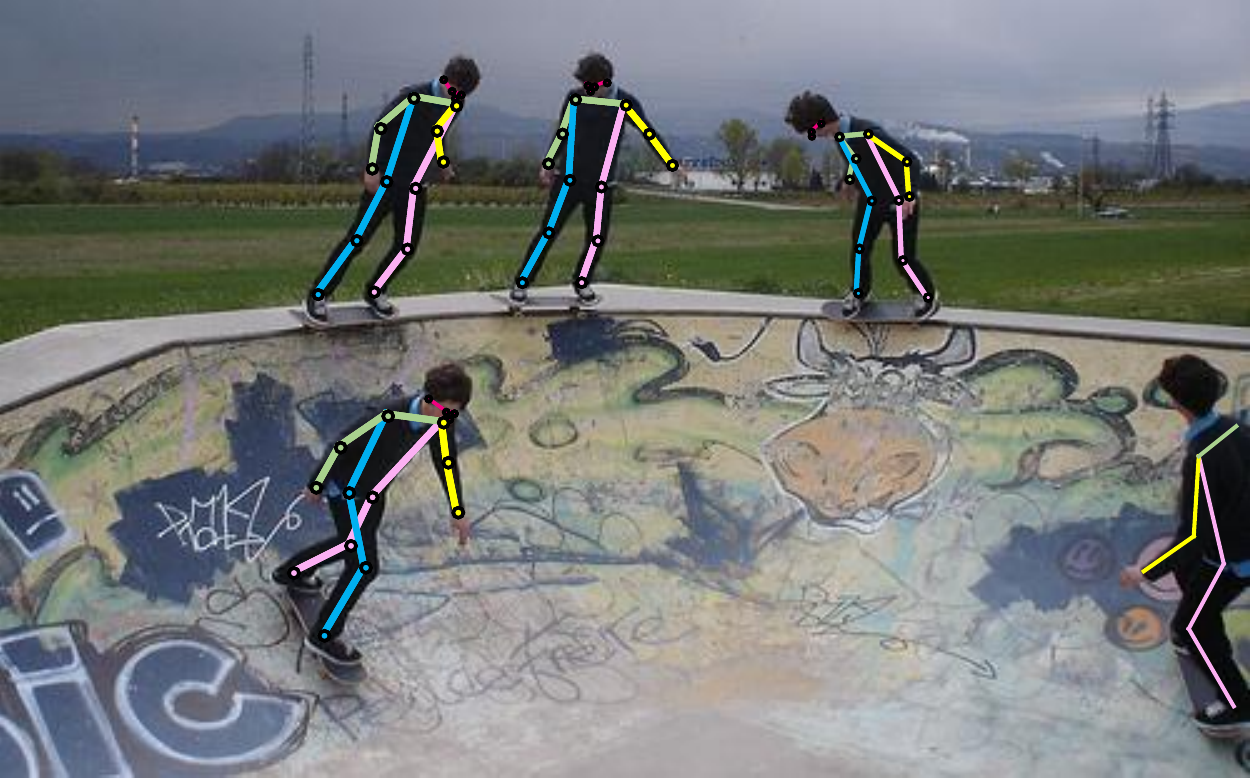}~\\
	~\includegraphics[width=0.24\textwidth,height = 0.183\textwidth]{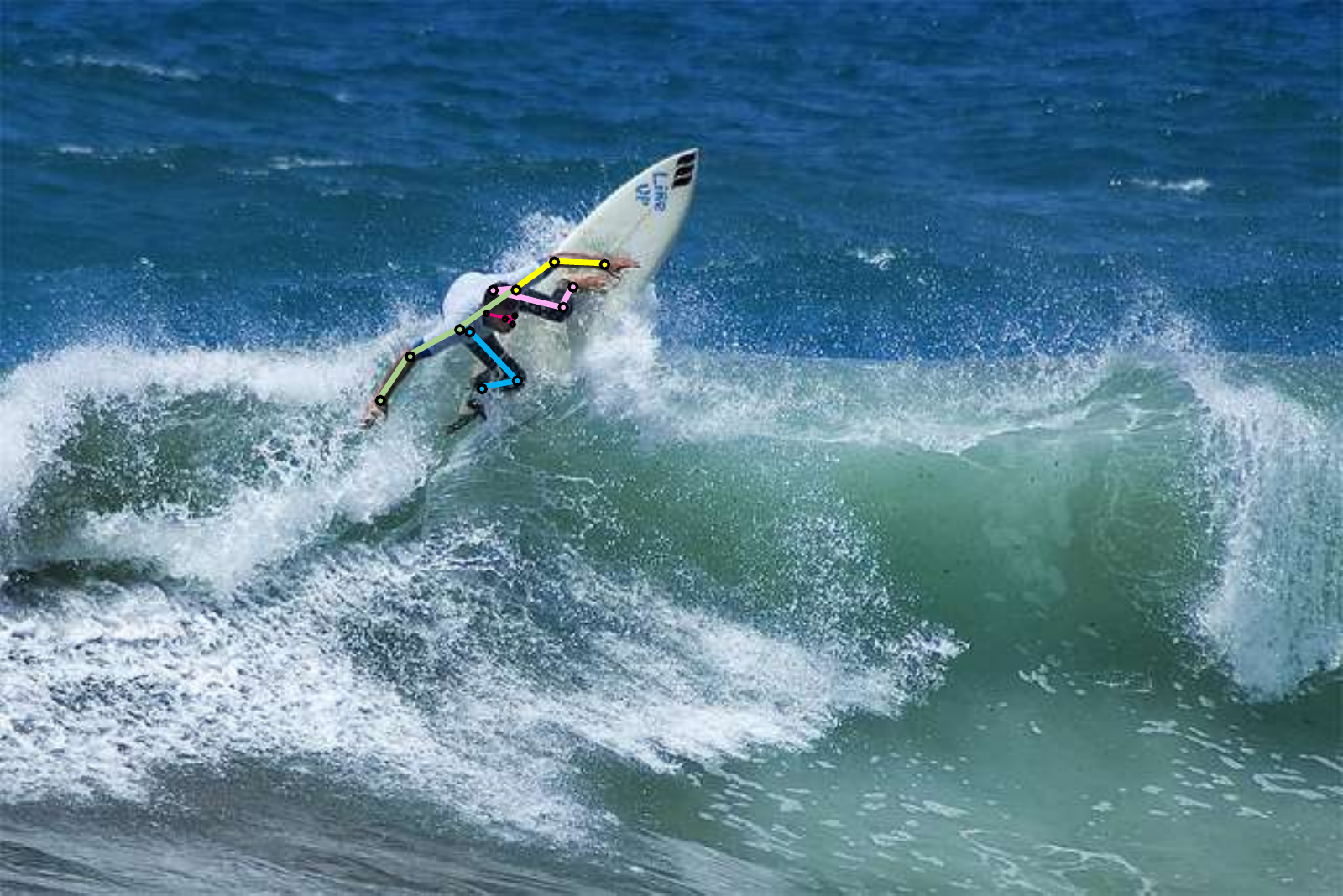}
	\includegraphics[width=0.24\textwidth,height = 0.183\textwidth]{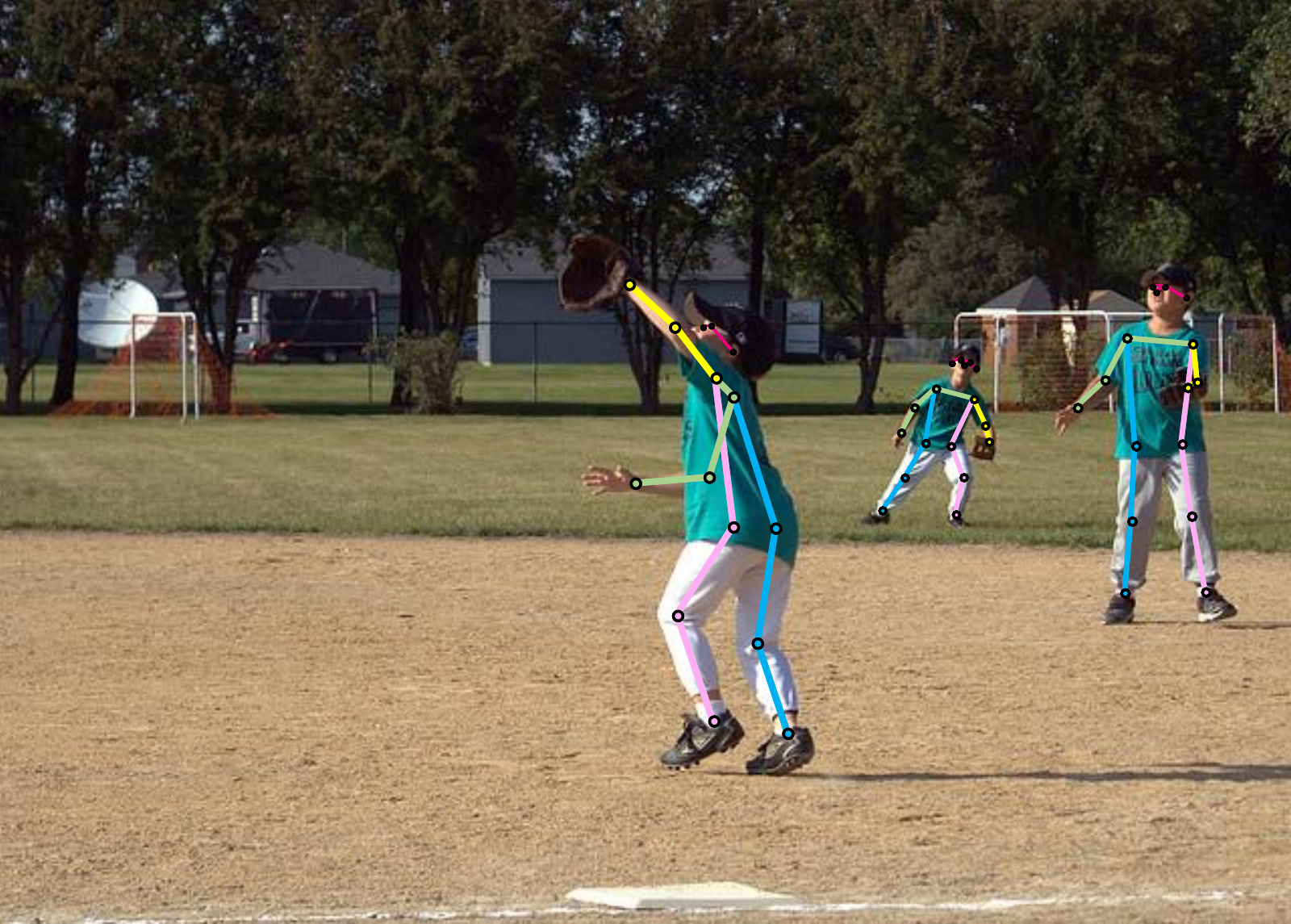}
 	\includegraphics[width=0.24\textwidth,height = 0.183\textwidth]{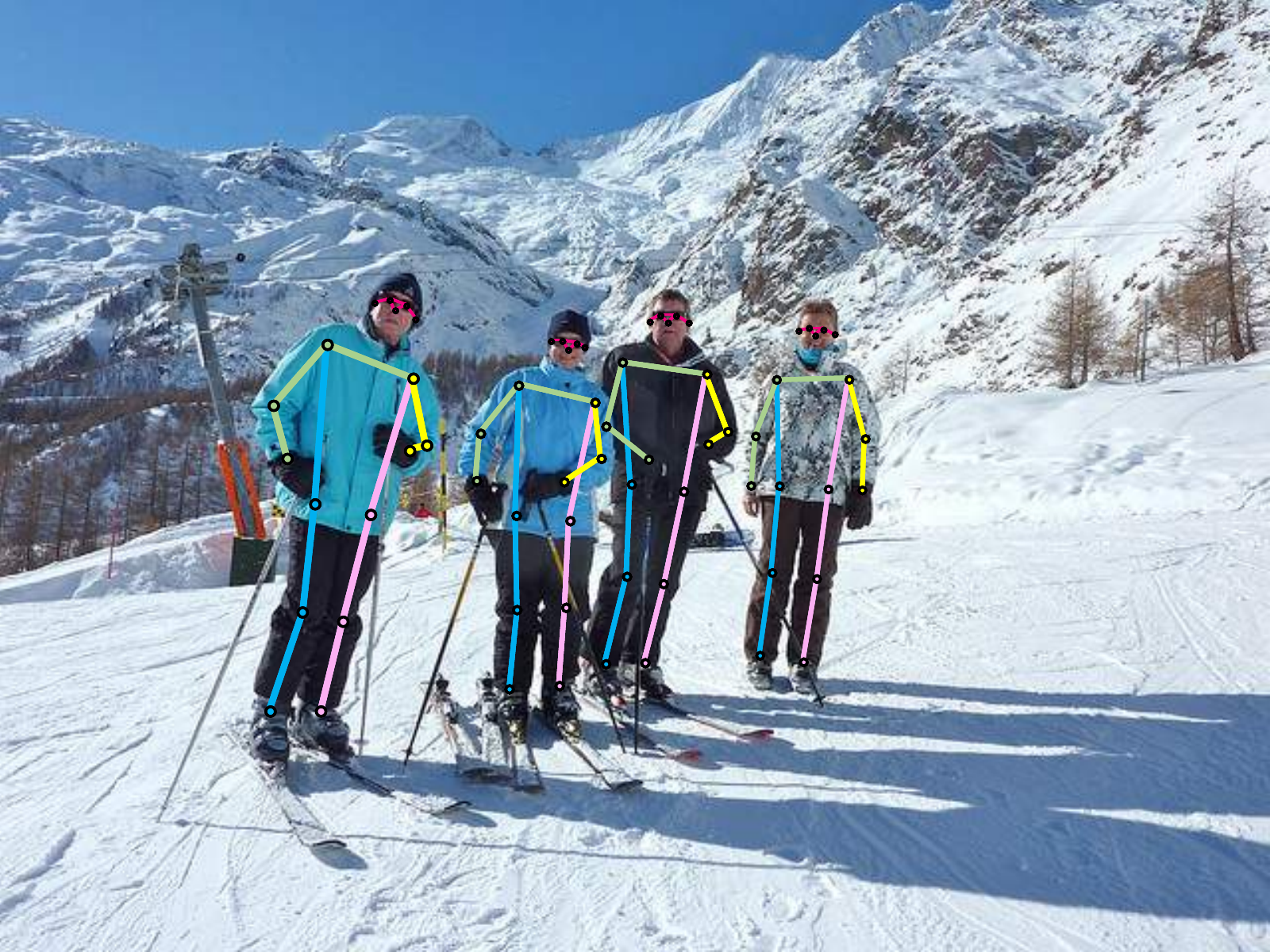}
 	\includegraphics[width=0.24\textwidth,height = 0.183\textwidth]{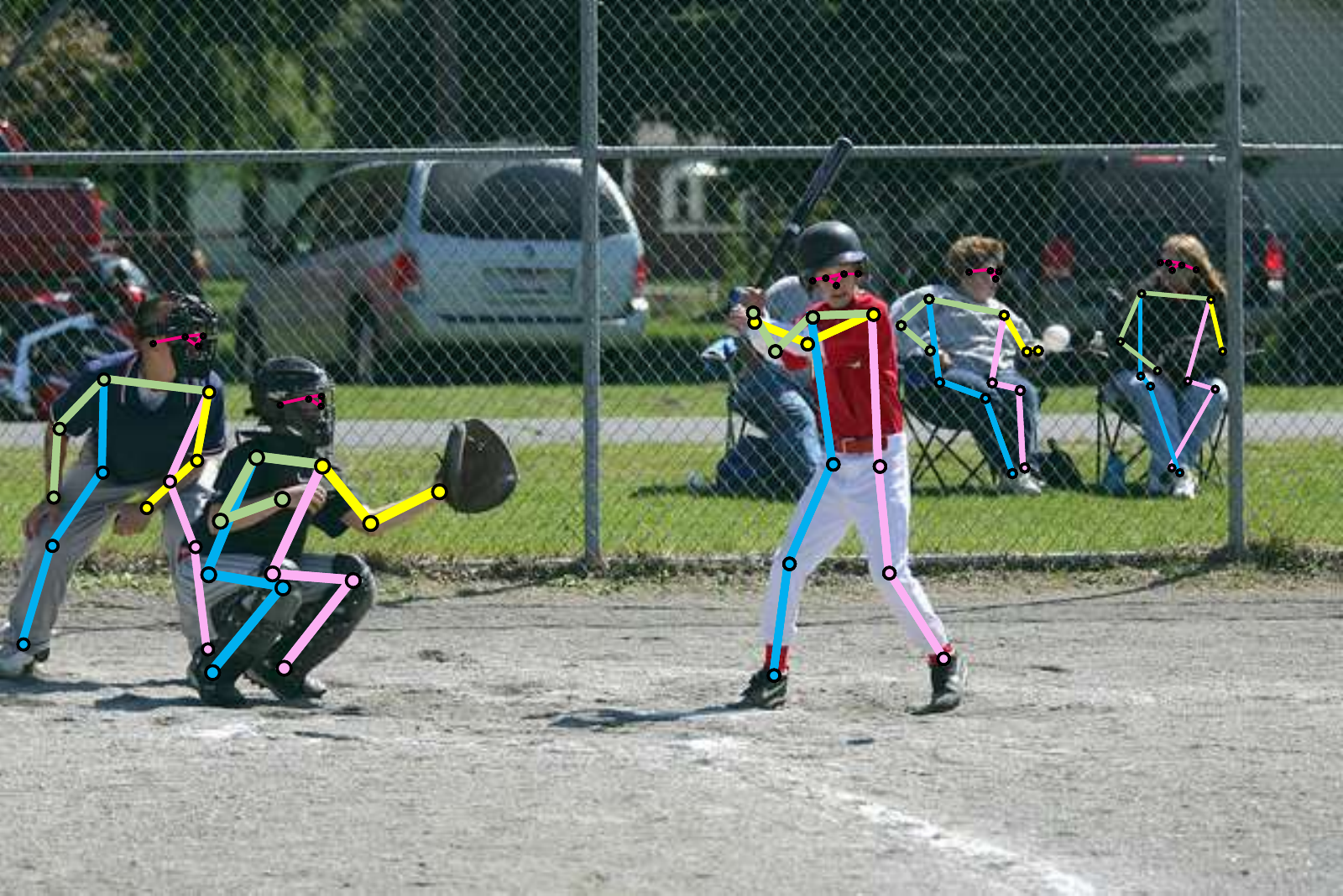}
\vspace{-.3cm}
	\caption{Qualitative results of some example images on COCO.}
	\label{fig:visual examples}
\end{figure*}

\begin{table}[t]
		\caption{Comparisons on the CrowdPose test set. $^*$ means using refinement. AE: Associative Embedding.}
		\vspace{-.3cm}
		\centering\setlength{\tabcolsep}{1pt}
		\label{table:crowdpose_test}
		\footnotesize
		\begin{tabular}{l|c|llllll}
			\hline
			Method & Input size &$\operatorname{AP}$ & $\operatorname{AP}^{50}$ & $\operatorname{AP}^{75}$ & $\operatorname{AP}^{E}$ & $\operatorname{AP}^{M}$ & $\operatorname{AP}^{H}$\\
			\hline
			\multicolumn{8}{c}{top-down methods}\\
			\hline
			Mask R-CNN~\cite{HeGDG17} & $-$ &$57.2$ & $83.5$&$60.3$&$69.4$&$57.9$&$45.8$\\
			AlphaPose~\cite{FangXTL17} & $-$ &$61.0$ & $81.3$&$66.0$&$71.2$&$61.4$&$51.1$\\
			SPPE$^*$\cite{li2018crowdpose} & $-$ &$66.0$ & $84.2$&$71.5$&$75.5$&$66.3$&$57.4$\\
			\hline
			\multicolumn{8}{c}{bottom-up methods: single-scale testing}\\
			\hline
			OpenPose~\cite{CaoSWS17} & $-$ &$-$ & $-$&$-$&$62.7$&$48.7$&$32.3$\\
			HrHRNet-W$48$ + AE ~\cite{cheng2019bottom} & $640$ & $65.9$ & $86.4$ & $70.6$ & $73.3$ & $66.5$ & $57.9$ \\
			\hline 
			Ours (HRNet-W$32$) & $512$ & $64.9$ & $84.5$ & $69.6$ & $72.7$ & $65.5$ & $56.1$\\
			Ours (HRNet-W$48$) & $640$ & $66.1$ & $84.6$ & $71.2$ & $73.4$ & $66.9$ & $57.1$\\
			Ours (HrHRNet-W$48$) & $640$ & $66.2$ & $84.9$ & $71.4$ & $73.6$ & $67.0$ & $57.6$\\
			\hline
			\multicolumn{8}{c}{bottom-up methods: multi-scale testing}\\
			\hline
			HrHRNet-W$48$ + AE~\cite{cheng2019bottom} & $640$ & $67.6$ & $87.4$ & $72.6$ & $75.8$ & $68.1$ & $58.9$ \\
			\hline
			Ours (HRNet-W$32$) & $512$ & $67.5$ & $86.1$ & $72.6$ & $75.5$ & $68.2$ & $58.2$\\
			Ours (HRNet-W$48$) & $640$ & $68.2$ & $85.7$ & $73.4$ & $75.9$ & $69.0$ & $58.9$\\
			Ours (HrHRNet-W$48$) & $640$ & $68.2$ & $86.2$ & $73.6$ & $75.8$ & $69.1$ & $59.1$\\
			\hline
			
		\end{tabular}
		\vspace{-.3cm}
	\end{table}

\subsection{CrowdPose}
    \noindent\textbf{Dataset.}
    We evaluate our approach on the CrowdPose \cite{li2018crowdpose} dataset
    that is more challenging and 
    includes many crowded scenes. The train set contains $10K$ images, the val set includes $2K$ images and the test set consists of $20K$ images. We train our models on the CrowdPose train and val sets and report the results on the test set as done in~\cite{cheng2019bottom}.
    
	\vspace{.1cm}
	\noindent\textbf{Evaluation metric.}
	The standard average precision based on Object Keypoint Similarity (OKS) which is the same as COCO are adopted as the evaluation metrics. The CrowdPose dataset is split into three crowding levels: easy, medium, hard.  
	We report the following metrics:
	$\operatorname{AP}$,
	$\operatorname{AP}^{50}$,
	$\operatorname{AP}^{75}$,
	$\operatorname{AP}^E$  for easy images,
	$\operatorname{AP}^M$  for medium images,
	$\operatorname{AP}^H$  for hard images.
	
	\vspace{.1cm}
	\noindent\textbf{Training and Testing.}
	The train and test methods follow COCO except the training epochs.
	We use the Adam optimizer~\cite{KingmaB14}.
	The base learning rate is set as $1\mathrm{e}{-3}$,
	and is dropped to $1\mathrm{e}{-4}$ and $1\mathrm{e}{-5}$ 
	at the $200$th and $260$th epochs, respectively.
	The training process is terminated within $300$ epochs.
	
	\vspace{.1cm}
	\noindent\textbf{Test set results.} 
	The results of our approach and other state-of-the-art methods 
	on the test set
	are showed in Table \ref{table:crowdpose_test}. 
	Our approach 
	with HRNet-W$48$ as the backbone achieves $66.1$ AP, 
	and outperforms the top-down methods, 
	leading to $8.9$ gain over Mask R-CNN~\cite{HeGDG17}, $5.1$ gain over AlphaPose~\cite{FangXTL17}. 
	With multi-scale testing, 
	our approach
	with HRNet-W$48$ as the backbone
	gets the best performance $68.2$ AP score, 
	much better than OpenPose~\cite{CaoSWS17}, $2.2$ gain over SPPE, and $0.6$ gain over HrHRNet~\cite{cheng2019bottom}.
	
\section{Conclusions}
	We present a baseline approach for
	improving bottom-up human pose estimation quality. 
	The success comes from: 
	exploiting the heatmaps
	for guiding pixel-wise keypoint regression,
	adaptive representation transformation (ART) for
	handling the diversity of human scales and rotations
	and better pixel-wise keypoint regression,
	heatmap tradeoff loss
	for improving heatmap estimation quality, 
	learning to scoring
	for promoting the pose candidates
	that are more likely to be true poses.

{\small
\bibliographystyle{ieee}
\bibliography{bib/hrnet2x,bib/BottomupPose}
}

\end{document}